\documentclass[journal]{IEEEtran}
\usepackage[T1]{fontenc}
\usepackage{amsmath,amssymb,graphicx,booktabs,tabularx,array,cite}
\usepackage[table]{xcolor}
\usepackage{xurl,capt-of,fancyhdr}
\usepackage[hidelinks]{hyperref}
\graphicspath{{figures/}}
\newcolumntype{Y}{>{\raggedright\arraybackslash}X}
\newcommand{\rev}[1]{#1}
\newenvironment{revised}{}{}

\definecolor{tableblue}{RGB}{21,61,89}
\hypersetup{
  pdftitle={EndoVLA-OOV: Language-Conditioned Out-of-View Target Acquisition in Robotic Endoscopy},
  pdfauthor={Chi Kit Ng, Yidong Zhang, Tianchun Wu, Zhiqing Tang, Hanlin Zhao, Siu Hing Lui, Tao Yang, Jingsong Lin, Huxin Gao, Yeung Trevor, Raymond Tang, Hongliang Ren},
  pdfsubject={Main manuscript and Supplementary Information},
  pdfkeywords={Robotic endoscopy, vision-language-action, out-of-view target acquisition}
}
\title{EndoVLA-OOV: Language-Conditioned Out-of-View Target Acquisition in Robotic Endoscopy}
\author{Chi~Kit~Ng, Yidong~Zhang, Tianchun~Wu, Zhiqing~Tang, Hanlin~Zhao, Siu~Hing~Lui,\\
Tao~Yang, Jingsong~Lin, Huxin~Gao, Yeung~Trevor, Raymond~Tang, and Hongliang~Ren%
\thanks{Chi Kit Ng, Yidong Zhang, and Tianchun Wu contributed equally to this work and share first authorship. (Corresponding author: Hongliang Ren.)}%
\thanks{Chi Kit Ng, Yidong Zhang, Tianchun Wu, Zhiqing Tang, Hanlin Zhao, Siu Hing Lui, Jingsong Lin, Huxin Gao, and Hongliang Ren are with the Department of Electronic Engineering, The Chinese University of Hong Kong, Hong Kong SAR, China (e-mail: hlren@ee.cuhk.edu.hk).}%
\thanks{Tao Yang is with The Sixth Affiliated Hospital of Sun Yat-sen University, Guangzhou, China.}%
\thanks{Yeung Trevor is with the Department of Surgery, Prince of Wales Hospital, The Chinese University of Hong Kong, Hong Kong SAR, China.}%
\thanks{Raymond Tang is with the Department of Medicine and Therapeutics, The Chinese University of Hong Kong, Hong Kong SAR, China.}}
\begin{document}
\bstctlcite{IEEErefcontrol}
\maketitle
\begin{abstract}
\rev{The scan-then-revisit workflow of upper gastrointestinal endoscopy requires bringing a previously observed suspicious region back into view for closer inspection. Acquiring such targets requires translating anatomical and appearance descriptions into camera motions from partial observations of a confined, deformable environment. We present EndoVLA-OOV, a language-conditioned vision-language-action policy for out-of-view target acquisition in robotic endoscopy. The policy adapts a pretrained backbone to continuous two-degree-of-freedom distal bending commands and combines initial, intermediate, and current images to guide search. To support policy learning, we provide a VLA dataset containing 1,050 episodes across six tasks and approximately 262k observation--action pairs. Physical evaluation spans wet-lens and low-light conditions, two unseen stomach phantoms, and cross-embodiment transfer to an unseen ex-vivo environment. In the cross-embodiment evaluation, the full-data policy achieves 93.3\% target acquisition success (28/30). In a controlled training-data comparison, expanding the corpus from 312 to 1,050 episodes increases acquisition success from 40.0\% to 90.0\% under the same model, training, and test configuration. Instruction and observation interventions further support the use of language and temporal context in the evaluated acquisition tasks. These results support language-guided endoscopic revisitation across the evaluated environments and robot embodiments and demonstrate the benefit of expanded task-specific demonstrations.}
\end{abstract}

\section{Introduction}
\begin{revised}
Upper gastrointestinal endoscopy combines systematic mucosal inspection with focused examination of suspicious regions. Screening protocols organize stomach coverage into anatomical stations and recorded views, providing a spatial framework for inspection and subsequent review~\cite{yao2013endoscopic,bravo2025gastrohun,veitch2015optimizing}. Quality standards emphasize careful examination and photographic documentation~\cite{beg2017quality,bisschops2016performance}, and a study of systematic image acquisition reported an association between more gastroscopy images and detection of clinically significant gastric lesions~\cite{lee2020taking}. These practices motivate robotic assistance that links a region of interest to the camera motion needed to inspect it. After an initial scan, an operator can specify an anatomical station and target appearance, and ask the robot to bring that target back into view for closer examination.

This scan-then-revisit workflow creates a specific control problem: language-conditioned acquisition of an initially out-of-view (OOV) target (Fig.~\ref{fig:workflow}). The instruction supplies a semantic goal, while the current image shows only a local portion of the stomach. Bending changes the visible folds, distances and illumination, and successive views reveal different parts of the target's surroundings. Successful acquisition therefore involves choosing camera motions from partial observations until the requested target becomes visible. Once acquired, the target supports image-space feedback for centering.

\begin{figure*}[t]
\centering
\includegraphics[width=\textwidth]{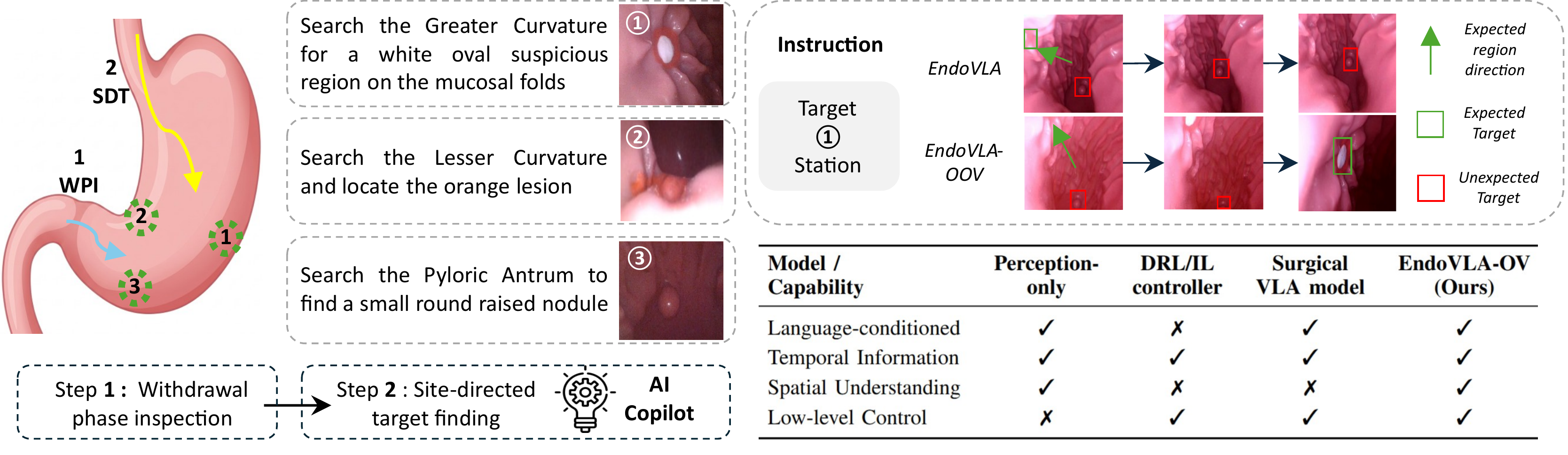}
\caption{Scan-then-revisit motivation and the language-conditioned distal-control task. The operator specifies target appearance and an anatomical station, and the policy generates bending actions. The schematic illustrates the intended interface; acquisition, centering and language dependence are evaluated separately.}
\label{fig:workflow}
\end{figure*}

Flexible endoscope operation combines insertion, axial shaft rotation and two distal bending degrees of freedom (DoFs). Robotic assistance has introduced motorized steering, lumen centralization and feedback control to support camera positioning~\cite{pullens2016colonoscopy,gafford2018fuzzy}. Visual servoing provides a general framework for converting visible feature errors into corrective motion~\cite{chaumette2006visual}. OOV acquisition extends the observation requirements of this interface: the policy uses the requested region, target appearance and previous observations to guide motion toward a target that becomes visible later in the trajectory. In this work, autonomous commands act on the two distal bending DoFs, while insertion and shaft rotation remain fixed throughout autonomous model trials. This defines a bounded action interface for studying acquisition in stomach phantoms and ex-vivo organs.

Vision-language-action (VLA) models connect semantic requests to robot commands. RT-1 learns language-conditioned actions from robot demonstrations~\cite{Brohan-RSS-23}; RT-2 transfers web-scale visual-language knowledge into control~\cite{pmlr-v229-zitkovich23a}; and OpenVLA demonstrates adaptation and generalization across downstream tasks~\cite{pmlr-v270-kim25c}. Cross-robot datasets and generalist policies extend this shared experience across tasks and embodiments~\cite{o2024open,Ghosh-RSS-24}. These pretrained representations provide visual and semantic priors beyond the available endoscopic demonstrations. This is relevant to low-light views with reduced contrast and wet-lens views with droplets and reflections, where target appearance shifts from the high-light, dry training condition. Broad pretraining therefore motivates our choice of a VLA backbone for language-guided OOV acquisition under appearance change. The physical evaluation measures transfer across appearance, organ geometry and robot hardware.

To support embodied learning in this setting, we provide a VLA dataset containing 1,050 episodes across six tasks and approximately 262k pairs.

EndoVLA-OOV adapts the pretrained $\pi_{0.5}$ action-chunk backbone~\cite{pmlr-v305-black25a} through low-rank updates~\cite{hu2022lora} and a continuous two-component action head. Its visual input combines an initial global view, an intermediate observation and the current frame. Action chunks represent a sequence of future bending commands, following a widely used approach to visuomotor sequence prediction~\cite{Zhao-RSS-23,black2024pi_0}. Together, the task interface, temporal observations and adaptation procedure form a domain-specific recipe for instruction-conditioned endoscopic acquisition.

We study transfer using two robotic embodiments: \textbf{R1} denotes the 6 mm platform used for demonstration collection and baseline evaluation, and \textbf{R2} denotes the 3.9 mm platform used for cross-embodiment evaluation. Experiments on R1 examine acquisition and composite finding-and-tracking (FT) under lighting and wetness changes and across two unseen phantoms and ex-vivo tissue. Experiments on R2 examine acquisition in an unseen ex-vivo environment and the effect of expanding the training corpus. Competing-target instructions and contextual-view interventions assess how language and temporal images guide the policy's search.

The transition from target acquisition to centering also motivates a clear account of feedback timing. A model that processes an input every $T_{\rm infer}$ seconds has perception rate $A=1/T_{\rm infer}$. A chunk of $H$ actions gives a theoretical maximum chunk-equivalent control rate $B=AH$ when all actions are consumed between successive processed observations; the chunk shares one input. Our common 300-input benchmark reports both quantities. This distinction connects action generation to the fresh-observation updates needed for precise centering and informs a future architecture that combines VLA acquisition with a low-latency visual controller. The Supplementary Information provides full data distributions, robot and annotation details, exact prompts, offline diagnostics and supplementary camera-control experiments.

The contributions of this work are:
\begin{itemize}
\item \textbf{An embodied endoscopic dataset:} We provide a VLA dataset linking language, endoscopic observations and recorded robot states to continuous two-DoF bending actions. Its 1,050 episodes span six tasks and approximately 262k pairs, covering both target tracking and OOV acquisition. This resource provides a shared data and control basis for developing and evaluating instruction-conditioned endoscopic policies.
\item \textbf{Temporal-context VLA control for OOV acquisition:} We introduce EndoVLA-OOV, which translates anatomical and appearance descriptions into continuous distal steering by conditioning a pretrained action-chunk policy on initial, intermediate and current views. Physical instruction and observation interventions support target selection through language and the use of contextual images during search. This formulation extends language-conditioned endoscopic assistance to targets beyond the current field of view.
\item \textbf{Transferable acquisition across endoscopic embodiments:} We demonstrate that a VLA policy adapted on R1 acquires language-specified targets on R2 in an unseen ex-vivo environment. Target acquisition reaches 93.3\% on R2, and a controlled training-data comparison raises acquisition from 40.0\% to 90.0\% with corpus expansion. These findings support reuse of the acquisition policy across the evaluated endoscopic platforms and identify demonstration coverage as a practical means of improving transfer.
\end{itemize}
\end{revised}

\section{Related Work}
\begin{revised}
\subsection{Robotic Endoscopy and Target Acquisition}
Robotic endoscopy supports camera positioning and instrument manipulation within a confined anatomical workspace. Endoluminal platforms provide coordinated actuation for interventions such as endoscopic submucosal dissection~\cite{chiu2025prospective,arezzo2024endoluminal}, while steering systems use motorized bending and image feedback to regulate the viewing direction~\cite{pullens2016colonoscopy,gafford2018fuzzy}. Visual servoing expresses a camera-positioning objective through measured feature errors and their relationship to motion~\cite{chaumette2006visual}. These approaches connect an observable image-space goal to corrective commands, making them relevant to lumen centralization and the centering of visible targets. For endoscopic revisitation, acquiring a requested target supplies the visual reference needed for this local feedback stage and links anatomical inspection to focused camera control.

Language-conditioned endoscopic policies extend this interface by allowing an operator to specify the feature to follow. EndoVLA~\cite{pmlr-v305-kit25a} combines a vision-language backbone with supervised and reinforcement fine-tuning for tracking tasks, predicting discrete directional or stop commands from the current image and prompt. This work establishes a direct connection between semantic target descriptions and endoscopic motion. EndoVLA-OOV addresses the preceding acquisition stage: the requested target begins outside the field of view, and the policy steers through successive observations until it becomes visible. Continuous bending actions and temporal visual context support this formulation. Table~\ref{tab:prior} summarizes the task, action, data and evaluation scope of the two approaches, positioning language-guided acquisition within the broader endoscopic control workflow.
\end{revised}

\begin{table}[t]
\caption{Scope relative to prior EndoVLA.}
\label{tab:prior}
\centering\footnotesize
\begin{tabularx}{\columnwidth}{@{}p{.19\columnwidth}YY@{}}
\toprule
Aspect & Prior EndoVLA~\cite{pmlr-v305-kit25a} & EndoVLA-OOV \\
\midrule
Task & In-view tracking & Initially OOV acquisition; composite FT \\
Actions & Four directions and stop & Continuous 2D chunks \\
Data & Approx.\ 6,000 pairs; 3 tracking tasks & 1,050 episodes; 6 tasks; 262k pairs \\
Evaluation & Tracking protocol & Second robot test; ex-vivo validation \\
\bottomrule
\end{tabularx}
\end{table}

\begin{revised}
\subsection{Language-Conditioned Control and Visual Context}
Language-conditioned control links semantic descriptions to spatially meaningful robot behavior. SayCan combines language-derived task proposals with robotic affordance estimates~\cite{brohan2023can}; CLIPort couples semantic representations with spatial action prediction~\cite{shridhar2022cliport}; and VIMA uses multimodal prompts to specify manipulation tasks~\cite{pmlr-v202-jiang23b}. PerAct represents language-conditioned actions in a voxel-based spatial representation~\cite{shridhar2023perceiver}, while VoxPoser composes language-driven three-dimensional value maps~\cite{pmlr-v229-huang23b}. Together, these methods connect what an instruction requests to where and how the robot acts through complementary representations of task intent, geometry and action. In endoscopy, anatomical-region and appearance descriptions similarly specify a target for camera motion. This connection motivates treating the instruction as part of the control problem, with acquisition measured for the requested target.

Semantic navigation extends language-conditioned behavior to goals encountered after a sequence of observations. LM-Nav grounds instruction landmarks through pretrained language, image-language and navigation models~\cite{shah2023lm}. VLFM combines observed geometry with a language-grounded value map to select exploration frontiers~\cite{yokoyama2024vlfm}. These approaches make observations gathered along a route useful for deciding where to search next. We draw on this relationship between semantic goals and accumulated visual evidence when formulating endoscopic OOV acquisition. Distal bending reveals different folds and surfaces, so the current image represents one part of the scene traversed during search. Temporal visual context offers a way to relate that local view to the preceding trajectory while preserving the requested target as the search objective.

\subsection{Pretrained VLA Policies and Cross-Embodiment Transfer}
Pretrained robot policies share representations across tasks and adapt them to a deployment interface. RT-1 learns language-conditioned control from robot demonstrations~\cite{Brohan-RSS-23}, while CLIP and PaLM-E establish complementary connections between visual observations, language and embodied inputs~\cite{radford2021learning,pmlr-v202-driess23a}. RT-2 incorporates robot actions into vision-language training to transfer semantic knowledge to control~\cite{pmlr-v229-zitkovich23a}, and OpenVLA provides an open pretrained policy with downstream adaptation~\cite{pmlr-v270-kim25c}. The $\pi_0$ family couples a vision-language backbone with continuous action generation, with $\pi_{0.5}$ emphasizing generalization through heterogeneous training~\cite{black2024pi_0,pmlr-v305-black25a}. Low-rank adaptation provides an efficient parameter-update mechanism for such specialization~\cite{hu2022lora}. These studies motivate adapting shared visual-language representations to endoscopic imagery and a compact distal-bending action space.

Training diversity and embodiment coverage determine the experience available to a shared policy. BridgeData V2 and DROID broaden manipulation demonstrations across tasks and collection settings~\cite{walke2023bridgedata,Khazatsky-RSS-24}; Open X-Embodiment aggregates experience across robots~\cite{o2024open}; and Octo studies generalist pretraining followed by adaptation to downstream observation and action spaces~\cite{Ghosh-RSS-24}. These resources connect transferable policy learning with variation in both visual conditions and robot interfaces. Endoscopic deployment introduces corresponding changes in tissue appearance, camera geometry and the response of the bending mechanism. This motivates an embodied dataset that links instructions and observations to recorded controls, together with evaluation of acquisition across environments and robotic platforms.

Continuous control also requires organizing predicted actions over time. Diffusion Policy generates action sequences using iterative denoising~\cite{chi2025diffusion}, ACT combines action chunks with temporal ensembling~\cite{Zhao-RSS-23}, and flow matching supplies the generative formulation used by flow-based VLA action experts~\cite{lipman2023flow,black2024pi_0,pmlr-v305-black25a}. Real-time chunking further addresses how prediction overlaps with execution under inference delay~\cite{black2026real}. These developments connect temporally coherent commands with the scheduling of observation updates. For endoscopic acquisition and subsequent centering, that relationship motivates reporting input-processing rate separately from generated-action throughput, linking the policy's sequence output to the feedback information available during control.
\end{revised}

\section{Policy Formulation and Adaptation}
\label{sec:method}
\subsection{Task and observation inputs}
The policy receives a language instruction $\mathbf{l}$ specifying target appearance and an anatomical region, and predicts distal bending actions. The phantom tasks associate a white oval target with the Greater Curvature (GC), a small round raised target with the Pyloric Antrum (PA), and an orange protruding target with the Lesser Curvature (LC). OOV episodes begin with the target absent from the current view. Finding precedes subsequent centering; these phases organize the evaluation of one policy through two task endpoints.

\begin{figure}[t]
\centering
\includegraphics[width=.88\columnwidth]{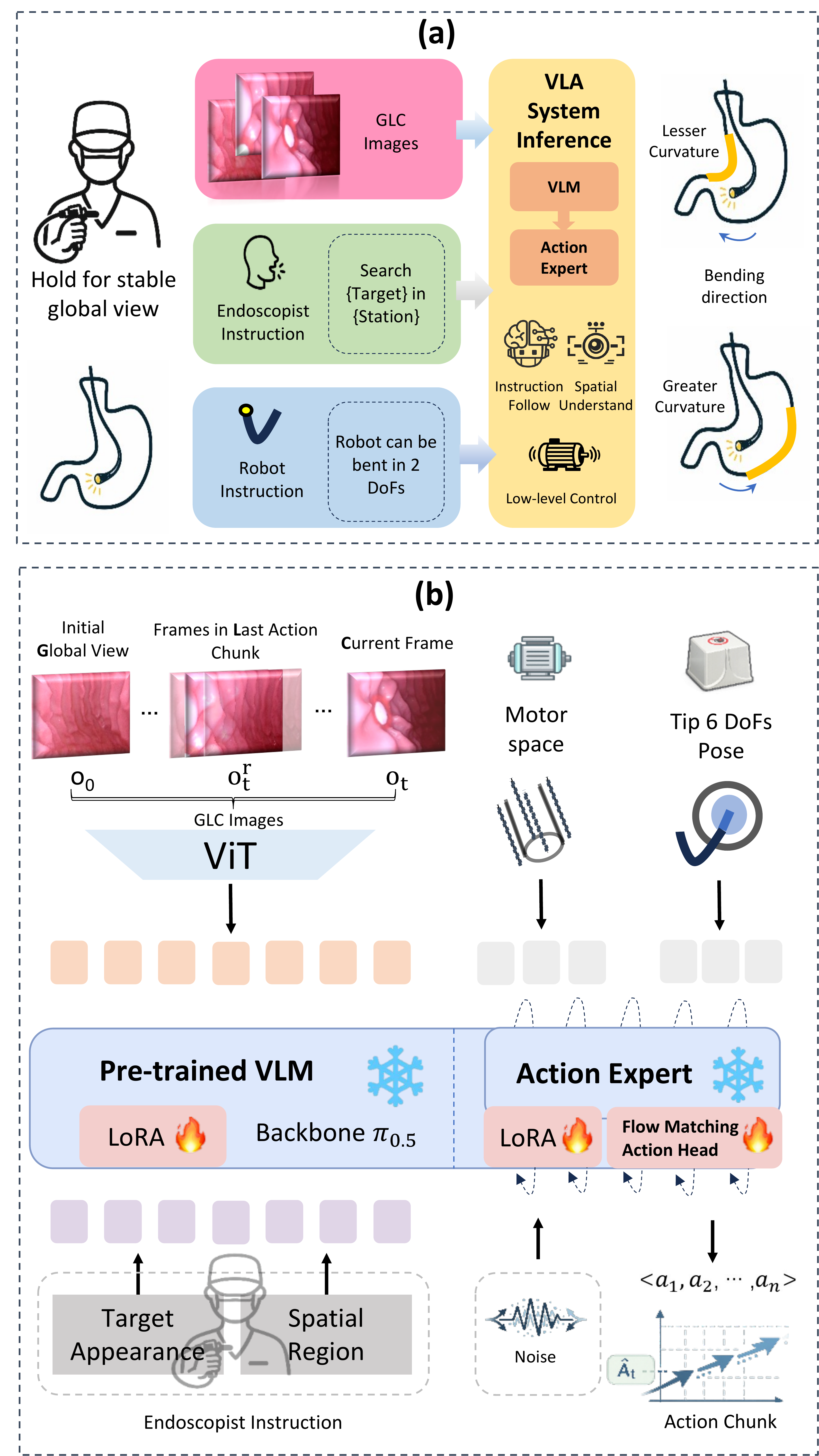}
\caption{Policy interface and adaptation architecture. The initial global, recent-history and current images provide temporal context to the vision-language model (VLM). A state encoder supplies continuous robot-state features directly to the action expert, which predicts two-dimensional bending-action chunks conditioned on the visual and language context.}
\label{fig:architecture}
\end{figure}

At decision time $t$, the visual bundle is
\begin{equation}
\mathbf{O}_t=(\mathbf{o}^{0},\mathbf{o}^{r}_t,\mathbf{o}_t),
\end{equation}
\rev{Here $\mathbf{o}^{0}$ is the global observation established by the operator at the beginning of the episode, providing an anatomical reference. The recent-history view $\mathbf{o}^{r}_t$ is one frame randomly sampled from images acquired during the executed portion of the preceding action chunk, providing recent motion context. The current view $\mathbf{o}_t$ is captured for the present decision and represents the target's current image location. A shared visual encoder maps the three images to visual tokens; the VLM processes these tokens with the language instruction. The view-mask experiment replaces one or both contextual images at inference while retaining the same trained model.}

The continuous robot state is
\begin{equation}
\mathbf{s}_t=[\mathbf{p}^{\mathrm{rel}}_t,\mathbf{p}^{\mathrm{abs}}_t,
\mathbf{q}_t,\mathbf{m}_t]\in\mathbb{R}^{12},
\label{eq:state}
\end{equation}
\rev{where $\mathbf{p}^{\mathrm{rel}}_t\in\mathbb{R}^3$ is tip displacement relative to the episode's configured origin, reducing sensitivity to coordinate-origin offsets; $\mathbf{p}^{\mathrm{abs}}_t\in\mathbb{R}^3$ is the measured absolute tip position, obtained from a localization system such as electromagnetic tracking; $\mathbf{q}_t\in\mathbb{R}^4$ is the orientation quaternion; and $\mathbf{m}_t\in\mathbb{R}^2$ contains the two bending-motor position proxies. The relative position is referenced to the episode origin.}

\rev{Each state component is normalized using training-data statistics. A small learned state encoder maps this continuous vector to a feature representation supplied directly to the action expert. The expert combines these features with the VLM's image--instruction context, a noisy action sequence and the flow-time parameter. This continuous-state pathway adapts the pretrained $\pi_{0.5}$ architecture to endoscope control; the $\pi_{0.5}$ baseline encodes discretized state values as language tokens (Supplementary Sec.~S2).}

\subsection{Continuous action interface}
\rev{Pretrained manipulation VLAs learn embodiment-specific proprioception and joint- or end-effector-space action representations~\cite{black2024pi_0,pmlr-v305-black25a}. Our endoscope combines tip pose and motor-position observations with two bending-motor velocity commands. We therefore couple LoRA with a learned state encoder, action-input projection and action-output MLP to adapt the observation-to-actuator mapping. Zero-padding and output-coordinate selection provide dimensional compatibility; the learned interfaces address the endoscope's state and action semantics.}

The policy maps observations, language and state to an action chunk,
\begin{equation}
\widehat{\mathbf{A}}_t=\pi_\theta(\mathbf{O}_t,\mathbf{l},\mathbf{s}_t)
\in\mathbb{R}^{H\times2},\qquad H=20.
\end{equation}
Each row commands the two bending-motor velocities. Per-axis 1st/99th-percentile normalization from dataset A connects training actions to motor commands (Supplementary Sec.~S2.2). The two-dimensional MLP replaces the pretrained high-dimensional output head. Command saturation and rate limits bound distal bending; insertion and shaft rotation stay fixed during autonomous trials.

\rev{The $H=20$ prediction covers one second at 20 Hz motor dispatch. Back-to-back queries use the latest observation; each completed prediction replaces the unexecuted remainder while commands are dispatched every 50 ms. Once a sequence is exhausted, its last velocity is held until the next prediction, including for $H=1$ baselines. Thus actual chunk consumption depends on prediction arrival. Section~\ref{sec:timing} defines the inference benchmark; Supplementary Sec.~S2 documents both robot interfaces.}

\subsection{Domain adaptation and objective}
We adapt the pretrained $\pi_{0.5}$ backbone using low-rank adapters in the vision-language and action-expert components while training the continuous-state encoder and remapped action head. The remaining backbone weights are frozen. \rev{LoRA represents an adapted weight as $W=W_0+UV$, where $W_0$ is frozen and $U,V$ are trainable low-rank factors~\cite{hu2022lora}. The trainable head converts the action-expert representation into the two bending coordinates. Ours-A and Ours-B share this parameter-update strategy; full-corpus training identifies the data used for adaptation.}

\rev{The state encoder is a $12\!\rightarrow\!256\!\rightarrow\!1024$ MLP that prepends one state token to 20 action-expert tokens. A $2\!\rightarrow\!1024$ projection embeds noisy actions, and a $1024\!\rightarrow\!256\!\rightarrow\!2$ MLP predicts their flow field. LoRA ranks are 16 for the language transformer and 32 for the action expert. Supplementary Secs.~S2.2--S2.4 specify the optimization settings, trainable parameters and attention connections.}

Following the flow-matching formulation~\cite{lipman2023flow,pmlr-v305-black25a}, sample Gaussian noise $\boldsymbol\epsilon$ and a flow time $\tau$, and form
\begin{align}
\mathbf{X}_\tau&=\tau\boldsymbol\epsilon+(1-\tau)\mathbf{A}_t,\\
\mathbf{U}_\tau&=\boldsymbol\epsilon-\mathbf{A}_t.
\end{align}
The action expert predicts $\mathbf{V}_\theta=f_\theta(\mathbf{O}_t,\mathbf{l},\mathbf{s}_t,\mathbf{X}_\tau,\tau)$ and minimizes
\begin{equation}
\mathcal{L}_{\mathrm{FM}}=\mathbb{E}\!\left[
\frac{1}{H}\sum_{i=1}^{H}\|\mathbf{V}_\theta[i]-\mathbf{U}_\tau[i]\|_2^2
\right].
\end{equation}
\rev{The loss trains the action expert to recover the conditional flow field over a complete bending-action chunk. At deployment, the generated sequence represents future commands under the supplied observation bundle and instruction. Ours-A and Ours-B use the same architecture and training configuration, differing in the training corpus; the comparison therefore evaluates the complete A-to-B corpus expansion under a common adaptation recipe.}

\section{Dataset and Experimental Setup}
\label{sec:setup}
\subsection{Robotic platforms and collection}
\begin{figure}[t]
\centering
\includegraphics[width=\columnwidth]{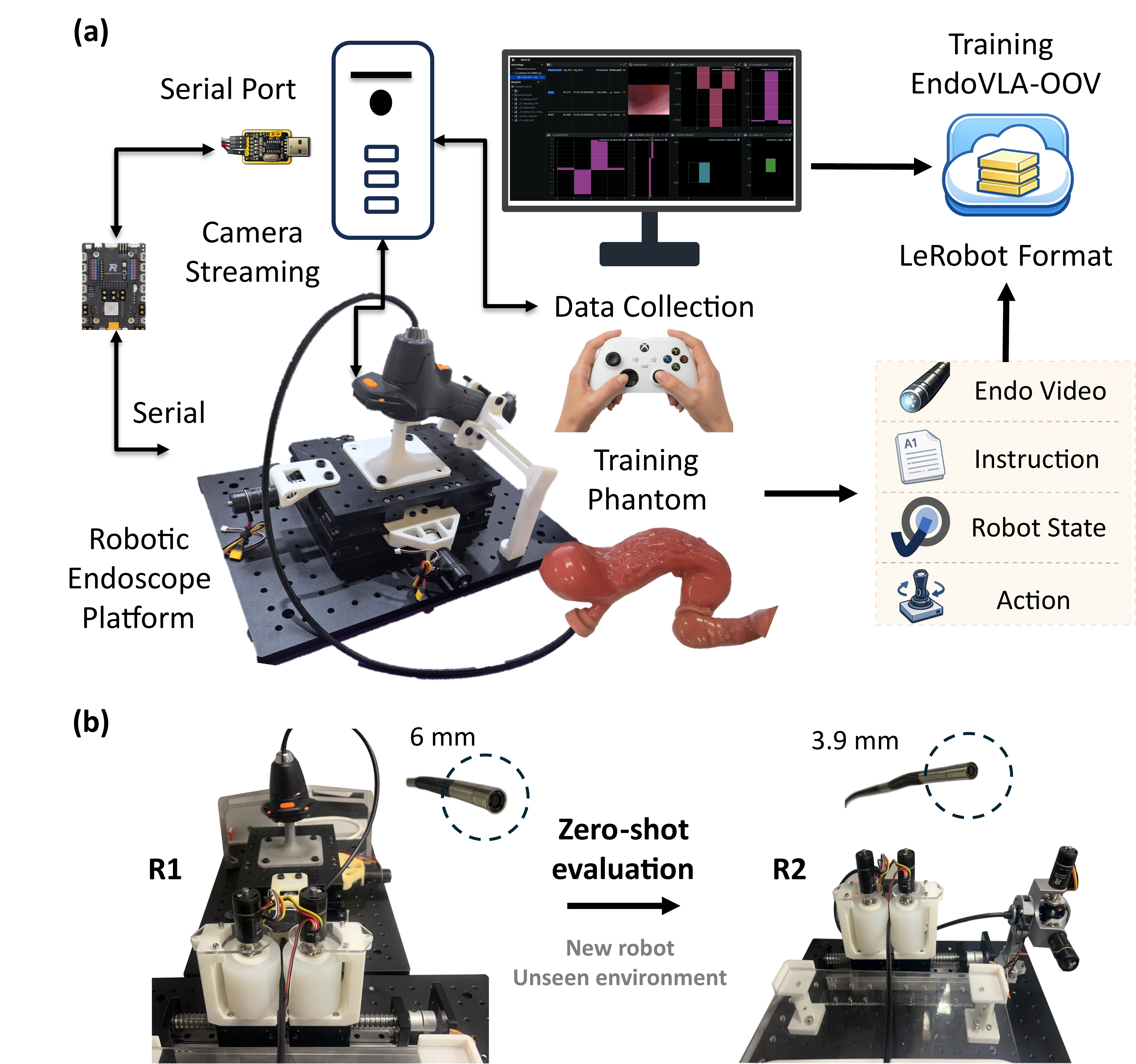}
\caption{\rev{Robotic platforms and the collection-to-evaluation workflow. (a) R1 teleoperation in the training phantom produces synchronized endoscopic images, instructions, recorded states and continuous bending commands, stored in LeRobot format for policy adaptation. (b) R1 uses a 6 mm endoscope and R2 a 3.9 mm endoscope. R1-trained policies are deployed on R2 in an unseen ex-vivo environment for zero-shot cross-embodiment evaluation.}}
\label{fig:collection}
\end{figure}

R1 motorizes the two bending axes of its endoscope using DJI RoboMaster M2006 P36 motors. \rev{An Xbox interface supports teleoperated demonstrations in the training phantom (Fig.~\ref{fig:collection}a). The collection host synchronizes endoscopic images, task instructions, recorded states and continuous bending commands into LeRobot-format episodes for policy adaptation.} The configured camera resolution is $640\times480$ and the recording rate is 20 Hz. \rev{Figure~\ref{fig:collection}b shows R1 with its 6 mm endoscope and R2 with its 3.9 mm endoscope. Policies trained from R1 demonstrations are directly evaluated on R2 in an unseen ex-vivo environment, constituting zero-shot transfer to the second robot. Both robots expose two distal bending commands. R2 applies a fixed factor of 0.3 to both joystick and autonomous bending-velocity commands to accommodate its greater bending sensitivity. Insertion and shaft rotation remain fixed throughout autonomous model trials, with the operator establishing the starting configuration before policy execution. Supplementary Fig.~S1 shows both platforms and the ex-vivo evaluation organs.}

\subsection{Tasks and model-specific training data}
Each episode contains tuples $(\mathbf{o}_t,\mathbf{s}_t,\mathbf{l},\mathbf{a}_t)$. T0--T2 begin with the target visible; search tasks A--C begin with it OOV. Demonstrations were collected under high-light, dry (HD) conditions. Table~\ref{tab:data} gives the full corpus and training subset A.

\begin{table*}[t]
\centering\footnotesize
\caption{Task design and training-data provenance. Ours-A and all VLA baselines are fine-tuned on subset A. B is the full corpus used by Ours-B. Dataset A/B, task A/B/C and organ exA--F are distinct identifiers.}
\label{tab:data}
\begin{tabularx}{\textwidth}{@{}lYrrrr@{}}
\toprule
Task & Training instruction & A episodes & A pairs & B episodes & B pairs \\
\midrule
T0 & Tracking white oval suspicious region. & 52 & 13,356 & 150 & 36,632 \\
T1 & Tracking small round raised nodule. & 52 & 10,667 & 150 & 52,843 \\
T2 & Tracking the orange protruding lesion. & 52 & 9,386 & 150 & 31,564 \\
A (GC) & Search the Greater Curvature for a white oval suspicious region on the mucosal folds. & 52 & 7,364 & 200 & 27,999 \\
B (PA) & Search the Pyloric Antrum to find a small round raised nodule. & 52 & 18,377 & 200 & 50,536 \\
C (LC) & Search the Lesser Curvature and locate the orange protruding lesion. & 52 & 20,160 & 200 & 62,040 \\
\midrule
Total & Six tasks & 312 & 79,310 & 1,050 & 261,614 \\
\bottomrule
\end{tabularx}
\end{table*}
\begin{figure*}[!t]
\centering
\includegraphics[width=.80\textwidth]{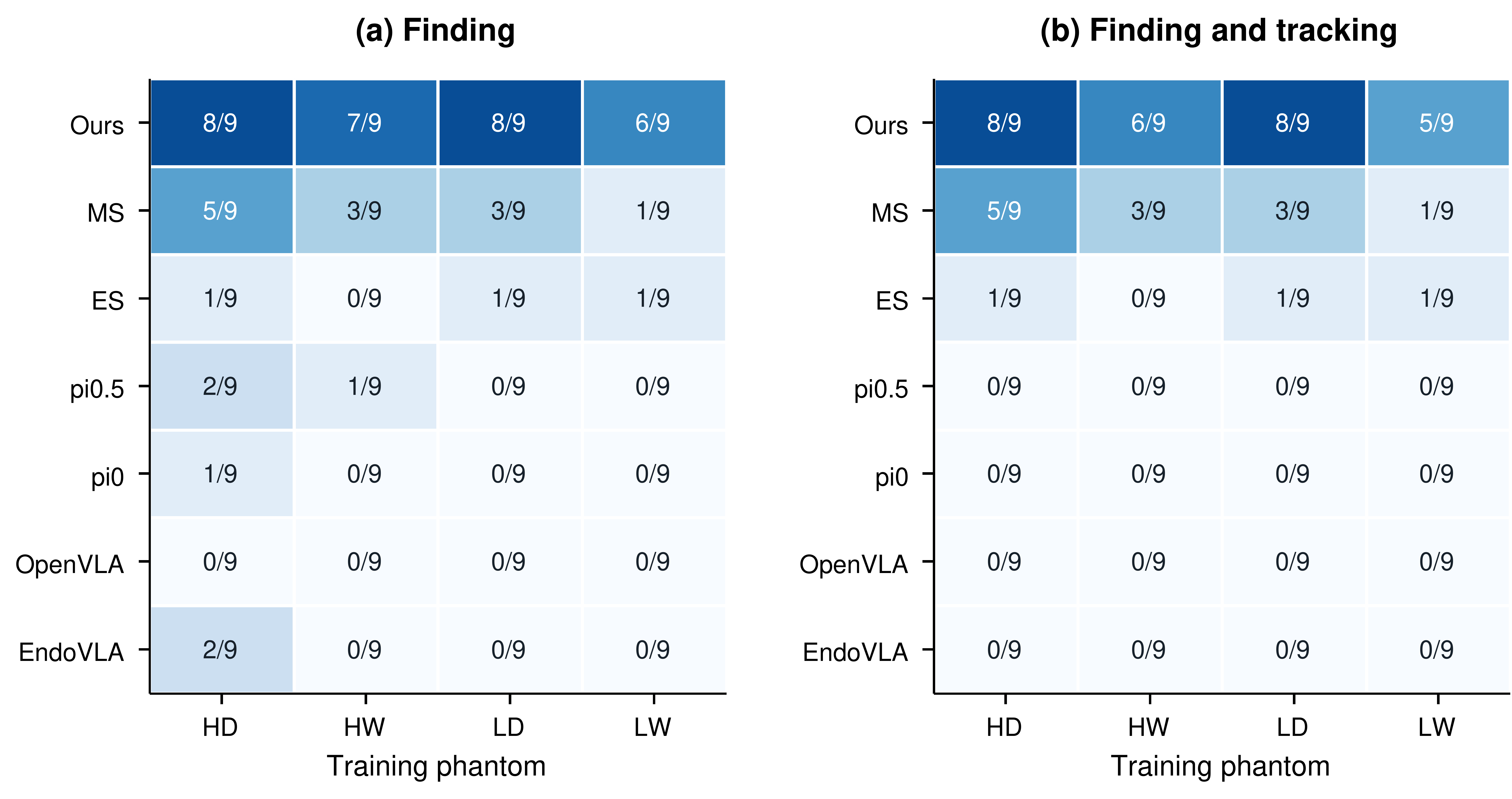}
\caption{System-level comparison on R1: (a) finding and (b) composite FT. Each entry is successes among nine trials. HD/HW/LD/LW denote high/low light and dry/wet conditions. Ours-A and all VLA baselines are fine-tuned on subset A. MS and ES denote one medical student and one engineering student, respectively. Supplementary Sec.~S8 specifies the R1 finding and FT protocol; Tables~S8a and S8b provide full counts and Wilson intervals.}
\label{fig:original}
\end{figure*}

\rev{Ours-A, EndoVLA, OpenVLA, $\pi_0$ and $\pi_{0.5}$ are fine-tuned on subset A. Ours-B is trained on the full corpus B and is used for exA/exB/exC and the training-phantom input interventions. The exD/exE/exF comparison evaluates Ours-A and Ours-B on R2, with the model identifiers specifying their respective training corpora. B increases both corpus size and the search-task episode fraction, from 50\% to 57.1\%. Its 1,050 episodes provide 3.37 times the episode count of A, and its 261,614 processed pairs provide 3.30 times the pair count. Processed pairs include temporally interpolated frames, so the corpus statistics describe recorded training entries with temporal dependence (Supplementary Sec.~S1).}

\subsection{Baseline and environment-transfer protocols}
The R1 evaluation is a \textbf{system-level comparison}: each adapted policy is assessed with its model-specific observation interface and training formulation on the common dataset A. It includes EndoVLA~\cite{pmlr-v305-kit25a}, OpenVLA-7B~\cite{pmlr-v270-kim25c}, $\pi_0$~\cite{black2024pi_0} and $\pi_{0.5}$~\cite{pmlr-v305-black25a}, adapted to the bending interface. \rev{The $\pi_0$ and $\pi_{0.5}$ baselines initialize from the official \texttt{pi0\_base} and \texttt{pi05\_base} checkpoints, respectively. Both receive the initial/global, recent-history and current images defined in Eq.~(1), the 12-dimensional state, and generate $H=20$ action chunks. Their standard 32-dimensional action interfaces are retained, with two-dimensional training actions zero-padded, flow loss evaluated only on the two physical action coordinates, and the first two output coordinates mapped to bending commands. LoRA adapts the attention and feed-forward layers of the PaliGemma language transformer and action-expert transformer with ranks 16 and 32, respectively; their pretrained non-adapter parameters are frozen. For $\pi_0$, the normalized continuous state is independently padded to width 32 and projected into the action expert. For $\pi_{0.5}$, normalized state values are discretized and encoded as language tokens. OpenVLA receives the current image and instruction, uses LoRA fine-tuning and predicts two action tokens per control step ($H=1$). The Qwen2.5-VL-based EndoVLA baseline receives the current image and instruction and uses supervised fine-tuning followed by group relative policy optimization (GRPO), with equally weighted L1-based action-alignment and output-format accuracy rewards (weight 1 each; summed reward in $[0,2]$). Every model receives 40,000 supervised/flow-matching updates with effective batch 32, using one RTX 5090 GPU, microbatch 4 and eight-step gradient accumulation. Dataset-A comparisons share uniform anchor sampling, seed 42 and the final fixed-budget checkpoint rule. EndoVLA receives 10,000 additional GRPO updates. Supplementary Sec.~S2 specifies learning rates, precision, normalization and trainable parameters.}

\rev{One medical student (MS) and one engineering student (ES) provide the two student-operator teleoperation comparisons.} Nine trials are reported per method and appearance condition: high-light dry (HD), high-light wet (HW), low-light dry (LD), and low-light wet (LW). Wet conditions introduce droplets on the lens and phantom surface. Environment-transfer evaluation uses R1 and Ours-A for FT trials on two unseen phantoms and ex-vivo anatomical regions. Supplementary Sec.~S7 reports camera-control experiments, including the random-action comparison.

\rev{Cross-embodiment evaluation on R2 examines acquisition with Ours-B and compares Ours-A/Ours-B with the remaining model, training and test conditions held fixed. Together, the baseline, environment-transfer and cross-embodiment experiments evaluate acquisition across appearance, anatomy and robot hardware. Supplementary Secs.~S4 and S8 provide the evaluation protocols and detailed trial outcomes.}

\subsection{Annotations, endpoints and uncertainty}
\rev{An external YOLOv8n detector proposes target bounding boxes. Our annotation workflow comprises manual boxes, detector fine-tuning, automatic sequence annotation and frame-by-frame human review; all inferred boxes and frame records were reviewed and corrected as needed. The detector supplies annotation proposals for evaluation; policy actions are generated from the observation, language and state interface in Sec.~\ref{sec:method}. Two private tracking corpora contain 544 episodes and 139,836 annotated frames, including 138,840 boxed frames. Supplementary Sec.~S3 reports the selected YOLO training/validation entries, internal mAP values and the review UI. The reported detector mAP values describe its internal validation split, and frame-by-frame human review supplies the annotation check for the evaluated sequences.}

\begin{revised}
R1 and R2 share the same Find criterion: at least 100 consecutive visible target frames within a 3,000-frame trial horizon, corresponding to five seconds within 150 seconds at 20 fps. After Find is satisfied, composite finding-and-tracking (FT) requires the target-box center to enter the central rectangle of width $0.5W$ and height $0.5H_{\rm img}$, with target visibility retained through the remaining trial. Center also requires prior Find success, followed by two uninterrupted seconds with the target-box center inside this same rectangle and the target continuously visible. The rectangle occupies 25\% of the image. The training-phantom input interventions use the same Find and Center definitions. Each rate uses all attempted trials as its denominator (Supplementary Secs.~S4 and S8).

We report $k/n$ counts and two-sided 95\% Wilson intervals under an independent-binomial model. The intervals describe trial-level counting uncertainty; physical trials are repeated within each organ and the interpretation remains specific to the evaluated specimens. We use descriptive comparisons of the counts, with specimen allocation and protocol details in the appendix. Reporting Find and Center separately identifies successful acquisition followed by incomplete centering.
\end{revised}

\section{Physical Evaluation}
\label{sec:results}
\subsection{Finding and find-and-track performance on R1}
Figure~\ref{fig:original} reports finding and composite FT results on R1. Ours-A finds the target in 8/9, 7/9, 8/9 and 6/9 trials under HD, HW, LD and LW, respectively, and completes FT in 8/9, 6/9, 8/9 and 5/9. These are the highest observed counts among the listed methods. MS achieves 5/9, 3/9, 3/9 and 1/9 in both endpoints; ES achieves 1/9, 0/9, 1/9 and 1/9. The other VLA methods have lower finding counts and composite FT counts of 0/9 in each appearance condition.

\rev{The observed counts support the adapted policy across all four tested appearance conditions. Ours-A retains 8/9 finding successes under both dry conditions, while the wet conditions yield 7/9 under high light and 6/9 under low light. Composite FT follows the same ordering, with 8/9 in the dry conditions and 6/9 and 5/9 in the wet conditions. This pattern identifies acquisition and subsequent centering in wet views as a useful focus for further improvement. Supplementary Sec.~S7 reports centering behavior and pixel-error analysis under the same appearance conditions.}

\begin{figure*}[t]
\centering
\includegraphics[width=\textwidth]{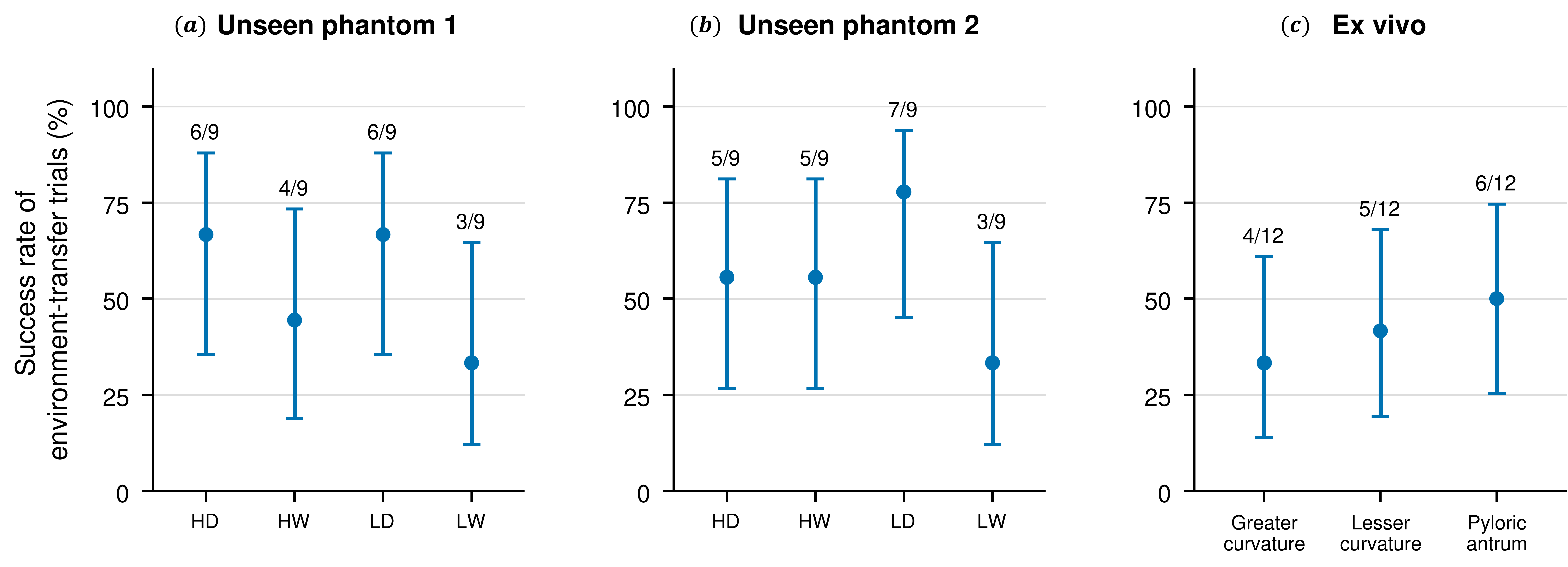}
\caption{Environment transfer of Ours-A on R1. Composite FT results use nine trials per unseen-phantom condition and twelve trials per ex-vivo anatomical region. Bars are 95\% Wilson intervals. The panel reports FT on R1; the cross-embodiment evaluation on R2 reports Find and Center separately.}
\label{fig:transfer}
\end{figure*}

\subsection{Environment transfer on R1 and illustrative rollouts}
\begin{figure}[t]
\centering
\includegraphics[width=\columnwidth]{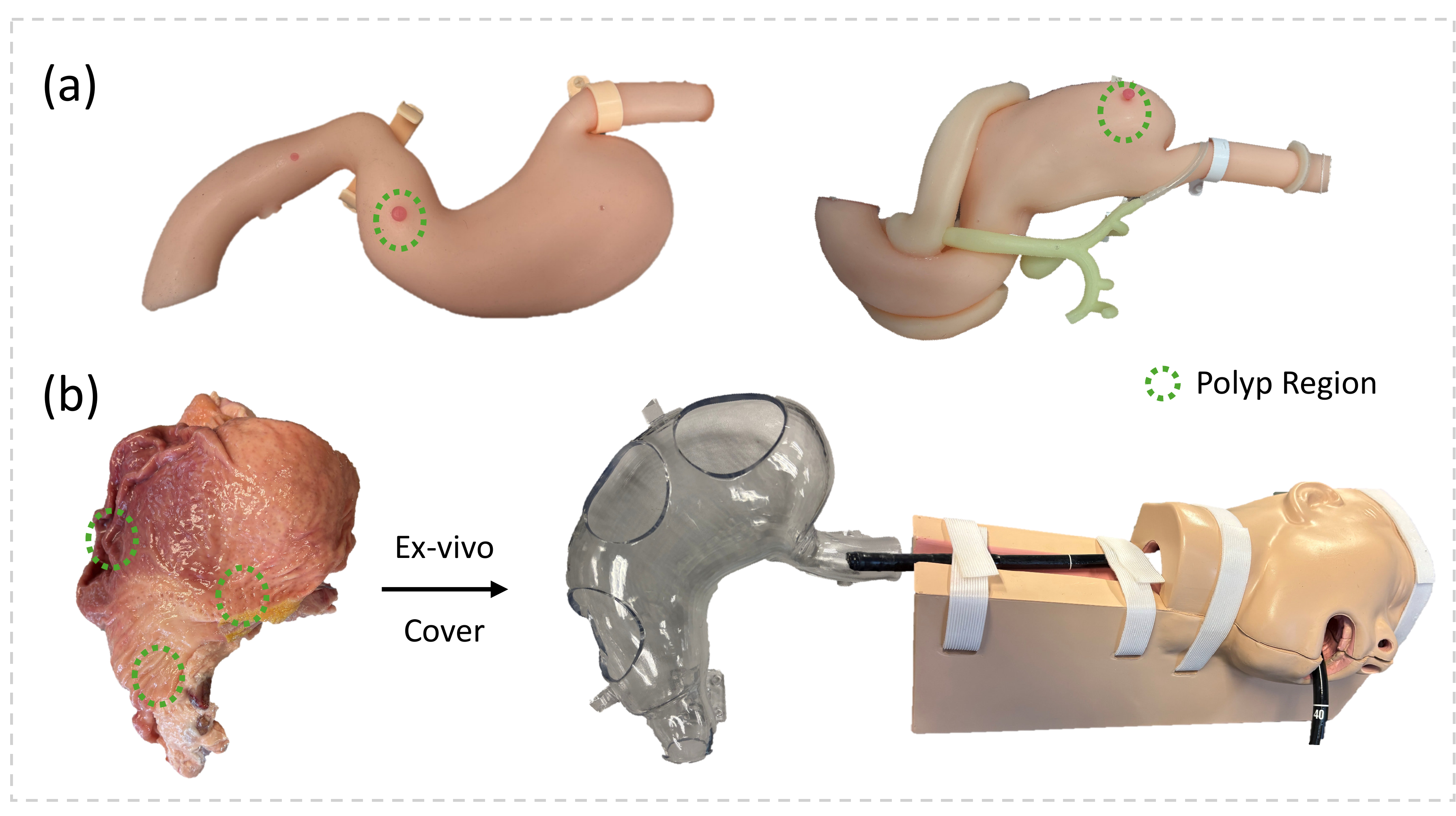}
\caption{Environment-transfer settings for Ours-A on R1: (a) two unseen stomach phantoms and (b) an ex-vivo preparation. The R2 evaluation uses organs exA--F.}
\label{fig:environments}
\end{figure}

\begin{figure*}[!t]
\centering
\includegraphics[width=.94\textwidth]{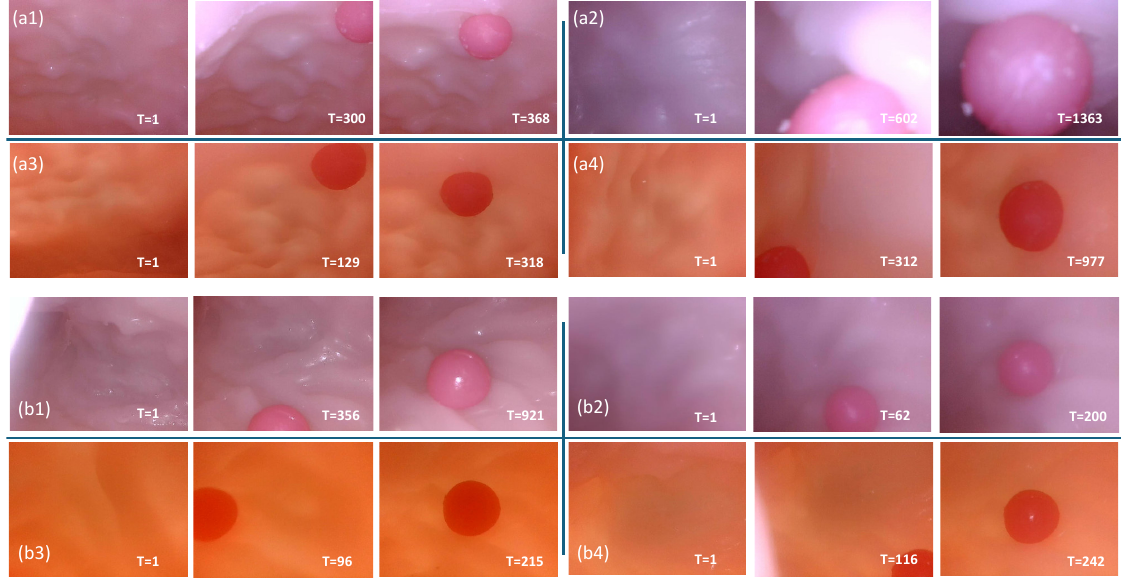}
\caption{Selected successful Ours-A FT rollouts on R1 in the two unseen phantoms. Each triplet shows an initial global view, an intermediate view and a final centered view. Panels (a1--a4) and (b1--b4) cover HD/HW/LD/LW on unseen phantoms 1 and 2. The panels illustrate completed search-and-center trajectories; Fig.~\ref{fig:transfer} reports condition-level success counts.}
\label{fig:phantomrollouts}
\end{figure*}

Figures~\ref{fig:environments} and~\ref{fig:phantomrollouts} show the transfer environments and successful Ours-A rollouts on R1. On unseen phantom 1, composite FT succeeds in 6/9, 4/9, 6/9 and 3/9 trials across HD/HW/LD/LW; the corresponding counts for phantom 2 are 5/9, 5/9, 7/9 and 3/9. Ex-vivo FT results on R1 are 4/12 for GC, 5/12 for LC and 6/12 for PA (Fig.~\ref{fig:transfer}; full tables in Supplementary Sec.~S8).

\rev{The transfer results demonstrate completed FT trajectories in each evaluated phantom condition and anatomical region. The lower wet-condition counts on the unseen phantoms identify the combined appearance and environment change as a demanding part of the tested setting. The rollouts show how a successful policy trajectory progresses from the initial view through search to centering. Supplementary Fig.~S7 illustrates this progression in selected training-phantom trials on R1, with corresponding aggregate counts in Fig.~\ref{fig:original}.}

\subsection{Cross-embodiment evaluation}
\label{sec:cross}
\begin{figure}[!t]
\centering
\includegraphics[width=\columnwidth]{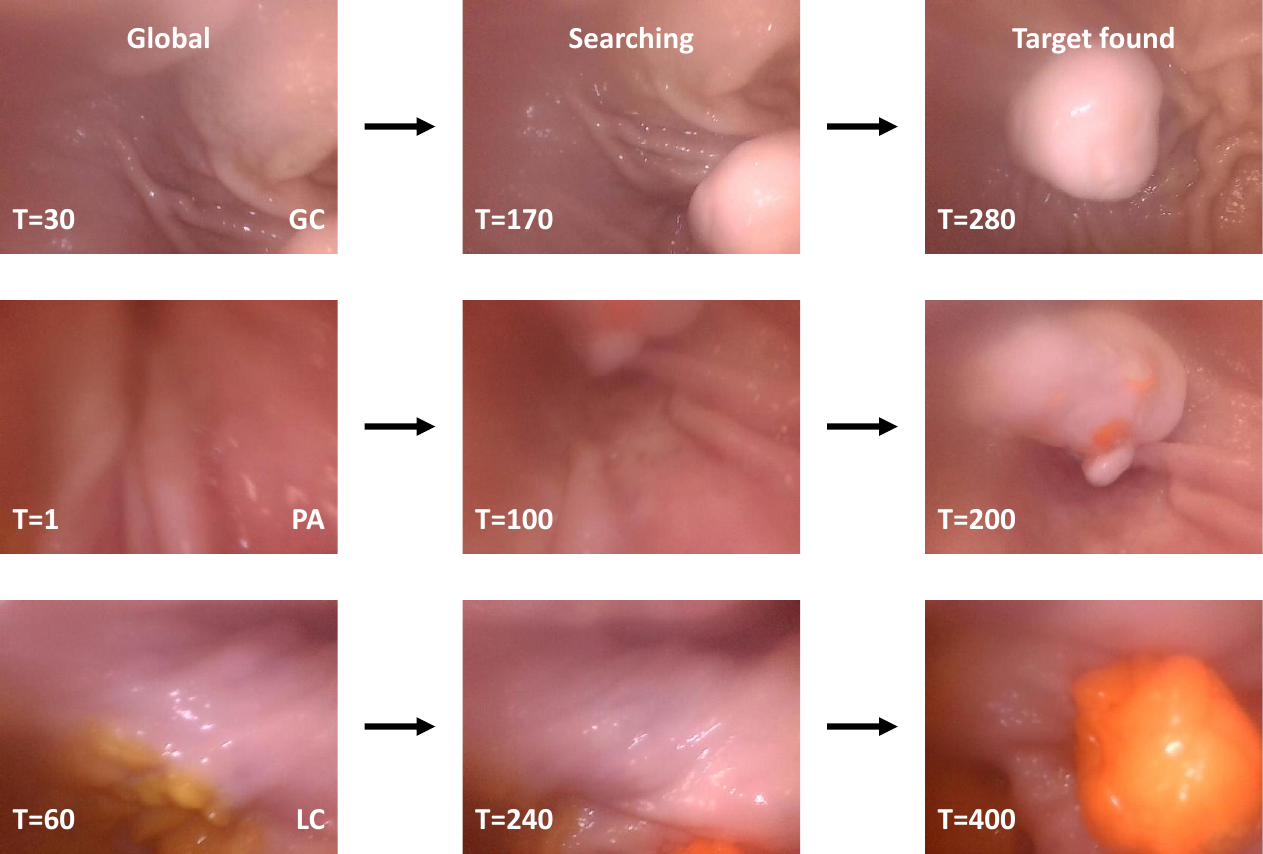}
\caption{\rev{Selected Ours-B target-acquisition rollouts on R2 in ex-vivo tissue. Rows show GC on exA, PA on exB and LC on exC, from top to bottom. Each sequence progresses from a global view through search to target acquisition. Figure~\ref{fig:abc} reports Find and Center outcomes across all standard trials.}}
\label{fig:crossexvivodemo}
\end{figure}

\begin{revised}
Cross-embodiment trials deploy the R1-trained policies on R2 (Fig.~\ref{fig:collection}b). The ex-vivo evaluation organs are denoted exA--F. exA/exB/exC support the standard acquisition, instruction and occlusion tests; exD/exE/exF support the training-data comparison. The standard prompt on exD matches exA, exE matches exB, and exF matches exC. Exact strings and intervention prompts are given in Supplementary Sec.~S4.

\subsubsection{Acquisition and strict centering with Ours-B}
Figure~\ref{fig:crossexvivodemo} illustrates target acquisition on the second robot across the three evaluated anatomical regions. The sequences connect successive camera motions with the requested target becoming visible.
Each standard group contains ten trials. Ours-B achieves Find/Center of 10/10 and 8/10 on exA, 9/10 and 6/10 on exB, and 9/10 and 6/10 on exC (Fig.~\ref{fig:abc}). Across the three standard groups, acquisition succeeds in 28/30 trials (93.3\%), compared with 20/30 (66.7\%) for strict centering. Of the 28 successful acquisitions, 20 also satisfy Center; eight acquired targets remain unsuccessful under the centering criterion.

\begin{figure}[!t]
\centering
\includegraphics[width=\columnwidth]{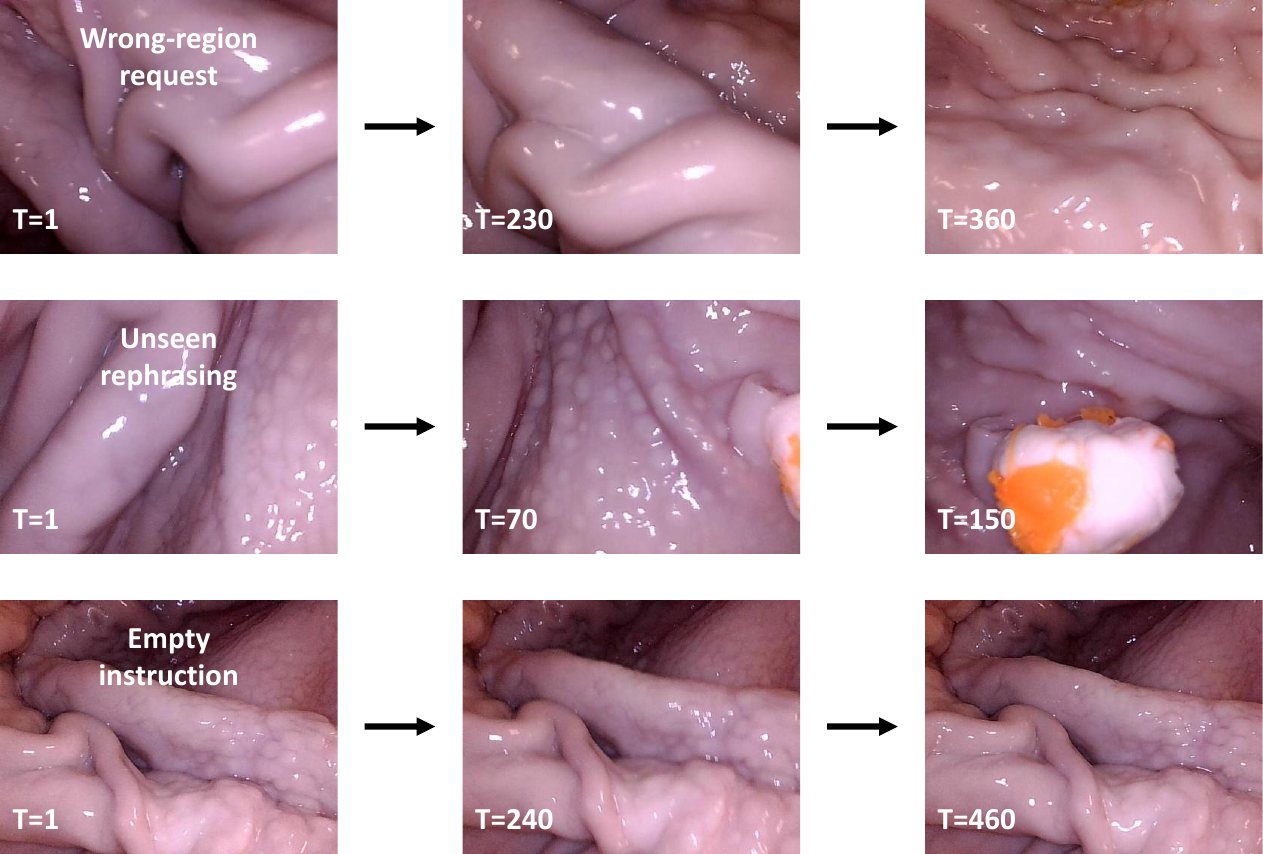}
\caption{\rev{Selected Ours-B rollouts on R2 under instruction interventions. Rows show a wrong-region request on exA, unseen correct wording on exB and an empty instruction on exC. The wrong-region rollout bends toward the target-free requested region; the unseen-wording rollout reaches the target, and the empty-instruction rollout continues searching. Figure~\ref{fig:abc} reports the corresponding ten-trial group outcomes.}}
\label{fig:instructiondemo}
\end{figure}

\subsubsection{Instruction conditions and occlusion}
The additional full-instruction group on exA yields Find 10/10 and Center 8/10. Unseen correct wording on exB gives 10/10 and 6/10; empty instructions on exC give 3/10 and 0/10. Wrong instructions on exA give 2/10 and 2/10 when scored against the reference target. This intervention changes the requested region and target description together, evaluating the effect of an alternative region--target request on acquisition of the reference target. Exact alternative instructions are listed in Supplementary Sec.~S4. The region specified by the wrong request contains no target. In the illustrated wrong-region rollout (Fig.~\ref{fig:instructiondemo}), the endoscope bends toward that region, leaving the reference target outside the view. On exC, the standard and empty-instruction groups yield 9/10 and 3/10. These organ-specific comparisons support a measurable dependence of acquisition on the supplied instruction.

After occlusion on exC, re-finding succeeds in 9/10 and Center in 6/10. This condition evaluates reacquisition after the target becomes occluded, with success scored through the same visibility and centering endpoints. The standard and additional conditions comprise eight groups of ten distinct videos (80 trials) on three organs, with 30/20/30 trials on exA/exB/exC. Each intervention is interpreted within its assigned organ; exact prompts, specimen assignments and the recovery-scoring scope are documented in Supplementary Sec.~S4.

\subsubsection{Controlled training-data comparison on exD--F}
We compare Ours-A against Ours-B on R2, changing only training data A/B while keeping architecture, training configuration and test conditions the same. Each model has ten trials per organ. Find improves from 3/10 to 10/10 on exD, 4/10 to 9/10 on exE and 5/10 to 8/10 on exF. Center improves from 1/10 to 7/10, 3/10 to 6/10 and 3/10 to 6/10, respectively (Fig.~\ref{fig:def}).

The totals are 12/30 versus 27/30 for Find (40.0\% versus 90.0\%) and 7/30 versus 19/30 for Center (23.3\% versus 63.3\%). Success improves on all three organs and both endpoints. Corpus expansion adds 15 Find successes and 12 Center successes across the 30 trials per model, corresponding to gains of 50.0 and 40.0 percentage points. The per-organ direction of improvement is consistent across all six endpoint comparisons. These observations support a practical benefit from the expanded training corpus under the fixed configuration, including its increased episode coverage and search-task share. Center remains below Find for Ours-B, retaining the acquisition--precision distinction observed on exA--C.
\end{revised}

\begin{figure}[t]
\centering
\includegraphics[width=\columnwidth]{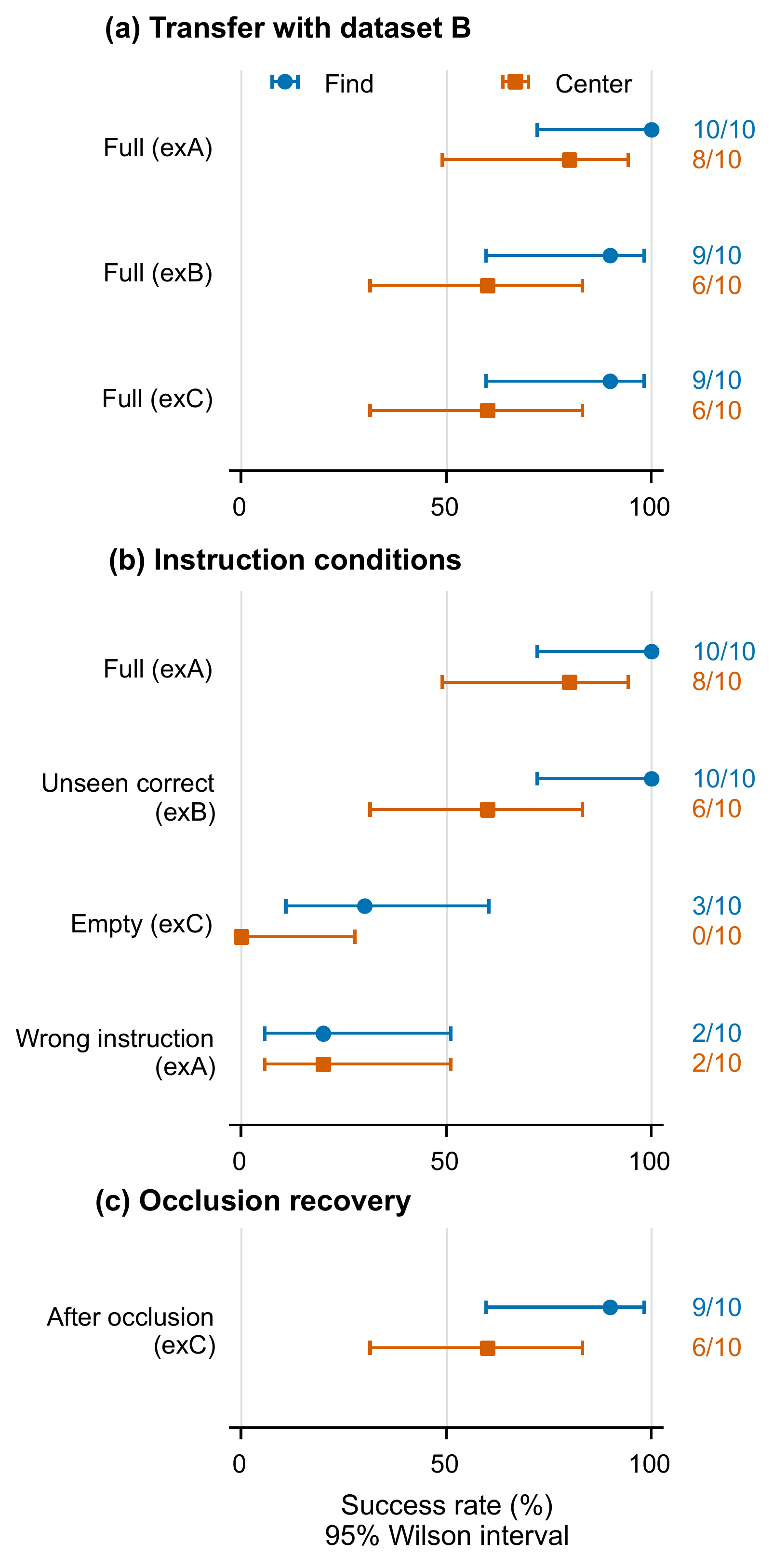}
\caption{\rev{Ours-B on R2: standard exA--C trials, instruction conditions and post-occlusion recovery. Each condition has ten trials; bars are 95\% Wilson intervals. Find and Center follow the criteria in Supplementary Sec.~S4. Standard exA and the separate full-instruction group each have Center 8/10. Each instruction condition is associated with the organ shown.}}
\label{fig:abc}
\end{figure}

\begin{figure}[t]
\centering
\includegraphics[width=\columnwidth]{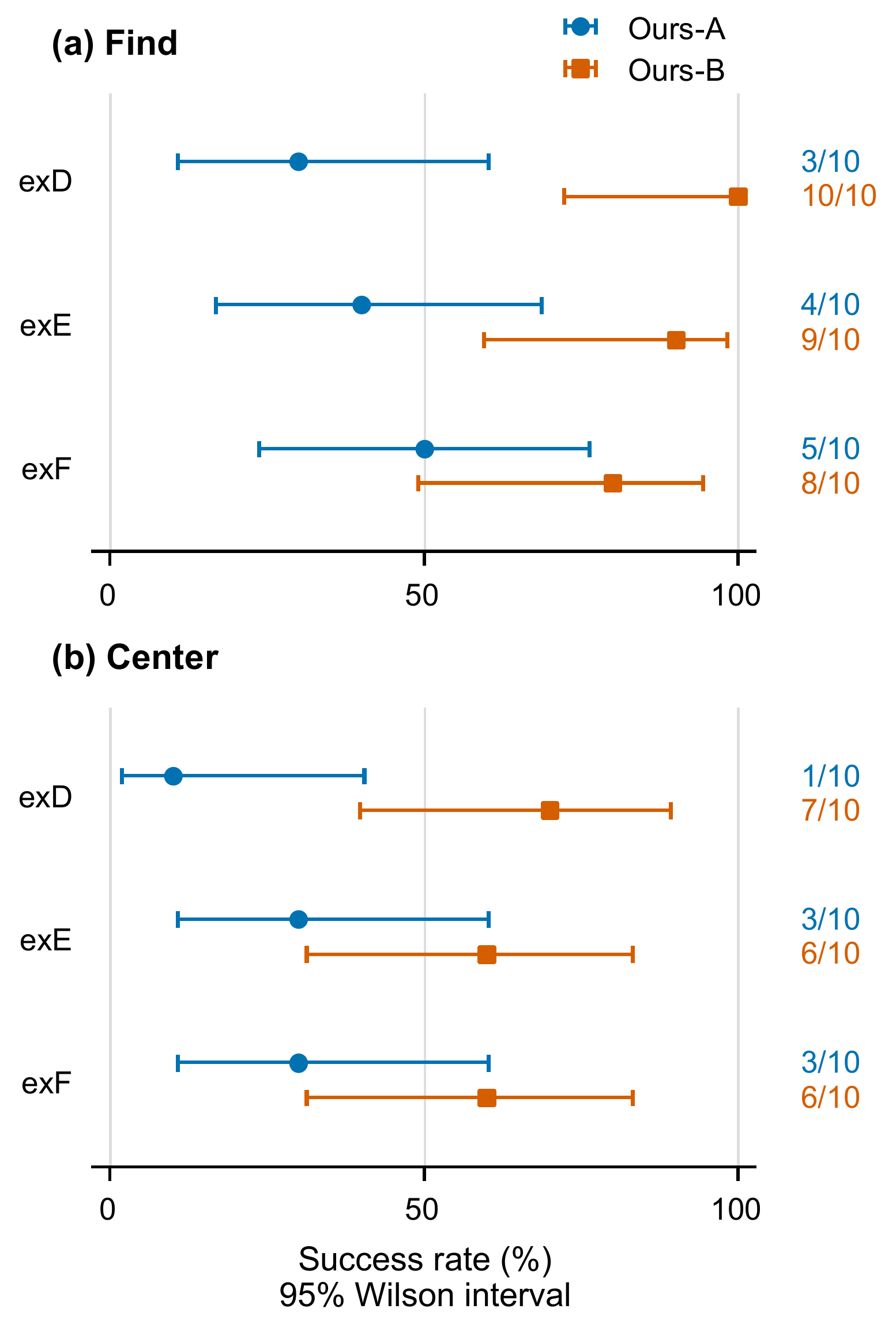}
\caption{\rev{Training-data comparison on R2 using organs exD--F. Ours-A uses 312 episodes and Ours-B uses 1,050 episodes; the other model, training and test conditions are unchanged. Each model has ten trials per organ. Bars are trial-level 95\% Wilson intervals; observations are repeated within each organ.}}
\label{fig:def}
\end{figure}

\subsection{Instruction and observation interventions in the training phantom}
\label{sec:interventions}
\begin{revised}
These physical input interventions use Ours-B in the training phantom. Two competing-target conditions request the GC white oval while ignoring the PA small round target, or request the PA target while ignoring the GC target. Each achieves Find 7/10 (70.0\%; Wilson 95\% interval 39.7--89.2\%). Success is acquisition of the requested target. The two reciprocal requests test selection between the specified region--appearance combinations, supporting instruction-conditioned target choice in this phantom.

For the instruction ``Search the Greater Curvature for a white oval suspicious region on the mucosal folds.'', we retain the same Ours-B checkpoint and mask zero, one or both contextual images (Table~\ref{tab:views}). Current-only input yields Find/Center of 6/10 and 5/10. Adding the recent view gives 7/10 and 6/10; adding the global view gives 7/10 and 7/10. All three views yield 9/10 for both endpoints, exceeding current-only input by 30 and 40 percentage points. These fixed-checkpoint results support temporal context for acquisition and centering. Each ten-trial group uses the common Find/Center criteria.
\end{revised}

\begin{table}[t]
\centering\footnotesize
\caption{\rev{Ours-B visual-context ablation. G/R/C: global/recent/current frames; omitted slots are masked. Entries: $k/n$ (\%; Wilson 95\% interval).}}
\label{tab:views}
\begin{tabularx}{\columnwidth}{@{}p{.25\columnwidth}YY@{}}
\toprule
Input (frames) & Find & Center for 2 s \\
\midrule
C (1) & 6/10 (60; 31.3--83.2) & 5/10 (50; 23.7--76.3) \\
R+C (2) & 7/10 (70; 39.7--89.2) & 6/10 (60; 31.3--83.2) \\
G+C (2) & 7/10 (70; 39.7--89.2) & 7/10 (70; 39.7--89.2) \\
G+R+C (3) & 9/10 (90; 59.6--98.2) & 9/10 (90; 59.6--98.2) \\
\bottomrule
\end{tabularx}
\end{table}

\subsection{Perception rate and maximum chunk-equivalent control rate}
\label{sec:timing}
\begin{revised}
We measure each model over 300 input-to-action-output inferences using common timing boundaries on an NVIDIA GeForce RTX 5090 GPU with 32 GB VRAM and an Intel Core Ultra 9 285K CPU. If $T_{\rm infer}$ is the mean time per input and $H$ the output chunk size, define
\begin{equation}
A=1/T_{\rm infer},\qquad B=H/T_{\rm infer}=A H.
\label{eq:rates}
\end{equation}
$A$ is the benchmark perception rate: the number of fresh inputs processed per second. $B$ is the theoretical maximum chunk-equivalent control rate, assuming all $H$ generated actions are consumed between successive processed observations, separated by $T_{\rm infer}$. Thus $1/B=T_{\rm infer}/H$ is an idealized action interval under this assumption. All actions in a chunk share the same input; $H=1$ gives $A=B$.

Table~\ref{tab:rates} reports $A=17.69$, 15.68 and 14.27 Hz for $\pi_0$, $\pi_{0.5}$ and Ours, respectively, with $H=20$. Ours has a mean inference time of 70.1 ms and $B=285.4$ actions/s. The physical motor interface instead dispatches one command every 50 ms (20 Hz). Processing capacity, observation scheduling and motor dispatch therefore describe separate parts of the control loop. The 20-step prediction covers a one-second horizon. During deployment, policy queries run back-to-back, with the next query starting immediately after the preceding prediction completes and using the latest available observation. Each completed chunk replaces the previous chunk's unexecuted remainder while motor dispatch continues at 20 Hz. Supplementary Sec.~S6 gives the timing interpretation and distinguishes inference latency from camera-to-actuator delay.
\end{revised}

\begin{table}[t]
\centering\footnotesize
\caption{\rev{300-input timing benchmark. $B$ is the theoretical maximum chunk-equivalent control rate; physical motor dispatch is 20 Hz.}}
\label{tab:rates}
\begin{tabular}{@{}lrrrr@{}}
\toprule
Model & $A$ (Hz) & $H$ & $B$ (Hz) & $1/A$ (ms) \\
\midrule
EndoVLA & 1.480 & 1 & 1.480 & 675.7 \\
OpenVLA-7B & 5.228 & 1 & 5.228 & 191.3 \\
$\pi_0$ & 17.69 & 20 & 353.8 & 56.5 \\
$\pi_{0.5}$ & 15.68 & 20 & 313.6 & 63.8 \\
Ours & 14.27 & 20 & 285.4 & 70.1 \\
\bottomrule
\end{tabular}
\end{table}

\section{Discussion}
\begin{revised}
\subsection{Motor-interface adaptation and embodiment transfer}
The system-level comparison distinguishes complete adaptation strategies. The $\pi_0$/$\pi_{0.5}$ baselines retain their 32-dimensional interfaces, pad the endoscopic state/actions, and select the first two output coordinates after LoRA adaptation; their pretrained interface parameters remain frozen. Ours jointly trains the 12-dimensional state encoder, two-dimensional action-input projection and action-output MLP with LoRA. Across R1 appearance conditions, $\pi_0$/$\pi_{0.5}$ achieve Find in 1/36 and 3/36 trials, compared with 29/36 for Ours-A. Under these configurations, the retained-interface, coordinate-extraction adaptations yield lower acquisition than the learned motor-control interface. This comparison supports the complete endoscopic adaptation; isolating individual interface contributions requires component ablations.

On R2 standard exA--C trials, Ours-B achieves Find in 28/30 attempts and Center in 20/30, supporting OOV acquisition across embodiment changes while identifying sustained centering as the more demanding endpoint.

On exD--F, replacing corpus A with B under the same model, training and test configuration raises Find from 12/30 to 27/30 and Center from 7/30 to 19/30. Gains on all three organs support expanded endoscopic search demonstrations.

\begin{figure}[t]
\centering
\includegraphics[width=\columnwidth]{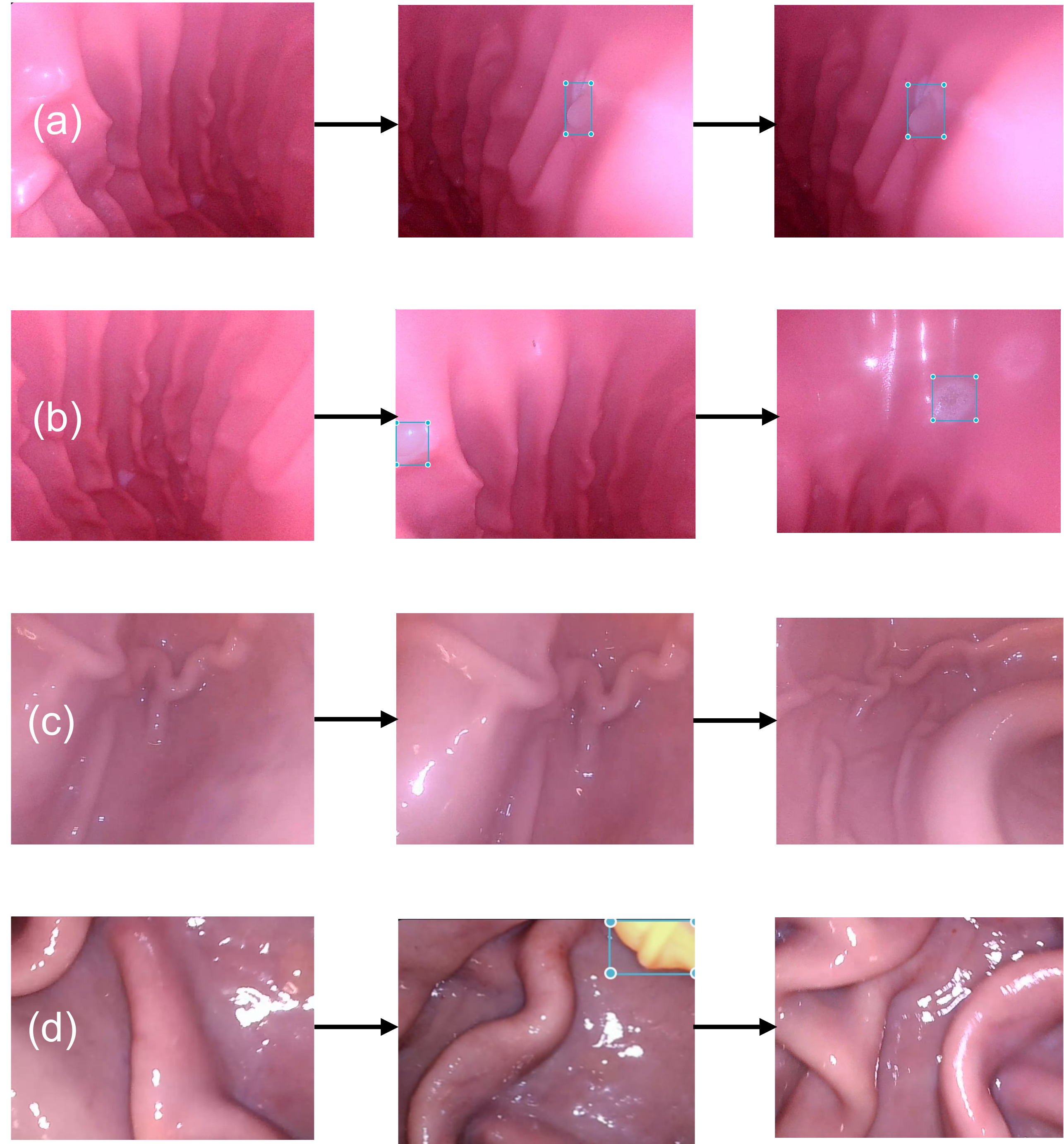}
\caption{Selected FT failures on R1 involving (a) a target-like water droplet, (b) a bubble, (c) search in an incorrect local region, and (d) incomplete reacquisition after partial approach. The examples illustrate visual discrimination and search-recovery challenges in the evaluated trajectories.}
\label{fig:failures}
\end{figure}

\subsection{Instruction use, context and observed failures}
Complete and alternative instructions on exA yield different reference-target acquisition counts; empty instructions reduce exC Find to 3/10. Reciprocal competing-target requests each achieve Find 7/10 in the training phantom. These observations support instruction-conditioned selection of the evaluated region--appearance combinations; Supplementary Sec.~S4 specifies prompts and organ assignments.

Current-only input achieves Find/Center of 6/10 and 5/10. Retaining either contextual image raises Find to 7/10 and Center to 6/10 or 7/10; all three views achieve 9/10 for both endpoints. This fixed-checkpoint comparison supports temporal context for acquisition and subsequent centering.

Figure~\ref{fig:failures} illustrates target-like droplets/bubbles and incomplete search or recovery. Condition-level analysis associates FT failures with wet appearance: across the three R1 phantoms, failures are 28/54 in wet views and 14/54 in dry views, with the same direction in each phantom. On standard exA--C trials, eight of ten Center failures occur after successful Find; this localizes the remaining difficulty primarily to post-acquisition control. Empty and wrong-region requests predominantly affect reference-target acquisition. These descriptive strata identify appearance robustness, instruction specificity and post-acquisition correction as distinct evaluation priorities (Supplementary Sec.~S8.4). Initial-pose thresholds and the causal contribution of each illustrated failure mode remain open questions. Supplementary Sec.~S5 reports fixed-noise instruction comparisons on 2,012 frames from 78 Ours-A validation episodes disjoint from A.

\subsection{Acquisition scope and visual-servo handover}
The contribution centers on language-guided acquisition of an OOV target. Once a target is visible, a reliable perception module, such as a YOLO detector, supplies localization for established visual-servo control~\cite{chaumette2006visual}. Tracking accuracy also depends on feedback latency, calibration and actuator response. The present comparison covers VLA systems and student operators. An independently tuned detector-plus-visual-servo baseline remains an evaluation limitation; the detector used here supplies human-reviewed scoring annotations. Pretraining motivates transfer across appearance changes, while direct evidence of an advantage over compact perception backbones remains a question for controlled comparison.

Ours processes 14.27 inputs/s (70.1 ms), with theoretical full-chunk throughput of 285.4 actions/s and actual motor dispatch at 20 Hz. Image age, target motion and actuator response jointly affect correction. ACT's temporal ensembling and real-time chunking illustrate coordination of successive predictions~\cite{Zhao-RSS-23,black2026real}, motivating separate assessment of fresh-observation processing and action throughput.

Future work will evaluate VLA-to-servo handover and compare lightweight learned, large pretrained and rule-based perception under a common controller in static, dynamic, low-light, wet-lens and occluded conditions. Localization accuracy, observation rate, centering error and retention will quantify tracking performance. Target-identity checks, confidence-based re-search and operator intervention will support handover evaluation.

\end{revised}

\section{Conclusion}
\rev{EndoVLA-OOV adapts a pretrained VLA policy to language-conditioned OOV target acquisition through continuous two-DoF endoscopic actions. The study provides a VLA dataset and links policy adaptation on R1 to environment-transfer and cross-embodiment evaluation on R2. The full-data policy achieves 28/30 acquisitions on standard exA--C trials, and corpus expansion increases Find from 12/30 to 27/30 on exD--F. Instruction and observation interventions support the use of language and temporal visual context in the evaluated tasks. Strict centering remains the more demanding endpoint, and separating perception rate from action-generation throughput defines a clear direction for improving feedback after acquisition.}

\section*{Acknowledgment}
\rev{ChatGPT (OpenAI) assisted with language revision and icon generation. The authors reviewed the resulting material and remain responsible for its scientific content.}

\bibliographystyle{IEEEtran}
\bibliography{references}

\input{supplement_format}
\thispagestyle{plain}
\pdfbookmark[0]{Supplementary Information}{supplement-start}
\begin{center}
{\Large\bfseries Supplementary Information\par}
\vspace{6pt}
{\large\bfseries EndoVLA-OOV: Language-Conditioned Out-of-View\\
Target Acquisition in Robotic Endoscopy\par}
\vspace{7pt}
{\small Chi Kit Ng, Yidong Zhang, Tianchun Wu, Zhiqing Tang, Hanlin Zhao, Siu Hing Lui\\
Tao Yang, Jingsong Lin, Huxin Gao, Yeung Trevor, Raymond Tang, and Hongliang Ren\par}
\end{center}
\begin{NoHyper}
\begingroup
\renewcommand{\thefootnote}{}
\footnotetext{Chi Kit Ng, Yidong Zhang, and Tianchun Wu contributed equally to this work and share first authorship. Corresponding author: Hongliang Ren (hlren@ee.cuhk.edu.hk).}
\endgroup
\end{NoHyper}
This supplement provides the data distributions, robot interfaces, annotation procedures and evaluation protocols supporting the main manuscript. It also reports offline action analyses and supplementary physical experiments, with each result interpreted in relation to language-guided target acquisition and subsequent camera control.

\section*{Contents}
\SIcontentsline{S1}{sec:si-data}{Training data and preprocessing}
\SIcontentsline{S2}{sec:si-robots}{Robotic embodiments, policy interfaces and adaptation}
\SIcontentsline{S3}{sec:si-annotation}{Detector-assisted annotation and human review}
\SIcontentsline{S4}{sec:si-protocols}{Instructions and evaluation protocols}
\SIcontentsline{S5}{sec:si-offline}{Offline instruction and action analyses}
\SIcontentsline{S6}{sec:si-timing}{Perception rate and maximum chunk-equivalent control rate}
\SIcontentsline{S7}{sec:si-control}{Supplementary camera-control experiments}
\SIcontentsline{S8}{sec:si-transfer}{Supplementary acquisition and transfer experiments}
\vspace{12pt}
\noindent\textbf{Notation and organization.}
R1 denotes the 6 mm demonstration-collection and evaluation robot; R2 denotes the 3.9 mm robot used for cross-embodiment evaluation. Datasets A and B identify the training subset and full corpus, respectively; exA--exF identify the R2 evaluation organs. Sections S1--S4 describe the experimental resources and protocols, S5--S6 examine action predictions and inference timing, and S7--S8 present supplementary physical results. Figure, table and equation numbers prefixed by S refer to this supplement. Citations use the reference list at the end of the main manuscript.
\clearpage

\section{Training data and preprocessing}\label{sec:si-data}
Ours-A and the EndoVLA, OpenVLA, $\pi_0$ and $\pi_{0.5}$ baselines are fine-tuned on dataset A, comprising 312 episodes and 79,310 observation--action pairs. Ours-B uses the full corpus B, comprising 1,050 episodes and 261,614 processed pairs. The exA--exC experiments on R2 and the training-phantom input interventions use Ours-B; the exD--exF experiments compare Ours-A and Ours-B under the same model, training and test configuration.

\setcounter{table}{0}
\renewcommand{\thetable}{S1}
\captionof{table}{Task distribution in training datasets A and B and the Ours-A evaluation set $E_A$. Pairs are processed observation--action pairs.}\label{tab:si-data}
{\footnotesize\setlength{\tabcolsep}{4pt}\renewcommand{\arraystretch}{1.18}\begin{center}\begin{tabularx}{\linewidth}{>{\raggedright\arraybackslash}X*{6}{>{\raggedleft\arraybackslash}X}}\toprule\rowcolor{tableblue}
\color{white}\bfseries Task & \color{white}\bfseries A episodes & \color{white}\bfseries A pairs & \color{white}\bfseries B episodes & \color{white}\bfseries B pairs & \color{white}\bfseries $E_A$ episodes & \color{white}\bfseries $E_A$ pairs \\\midrule
T0 & 52 & 13,356 & 150 & 36,632 & 13 & 2,488 \\
T1 & 52 & 10,667 & 150 & 52,843 & 13 & 2,357 \\
T2 & 52 & 9,386 & 150 & 31,564 & 13 & 2,893 \\
A: GC & 52 & 7,364 & 200 & 27,999 & 13 & 1,691 \\
B: PA & 52 & 18,377 & 200 & 50,536 & 13 & 2,711 \\
C: LC & 52 & 20,160 & 200 & 62,040 & 13 & 3,674 \\\midrule
\textbf{Total} & \textbf{312} & \textbf{79,310} & \textbf{1,050} & \textbf{261,614} & \textbf{78} & \textbf{15,814} \\
\bottomrule\end{tabularx}\end{center}}

Table~\ref{tab:si-data} shows that A contains 52 episodes per task. Corpus B contains 150 episodes per tracking task and 200 per search task, increasing episode count by 3.37 times and processed-pair count by 3.30 times. The search-task share increases from 50.0\% to 57.1\%. The controlled corpus comparison therefore evaluates the combined benefit of greater demonstration coverage and the expanded search-task allocation.

T0/T1/T2 involve initially visible targets. Search tasks A, B and C refer to the Greater Curvature, Pyloric Antrum and Lesser Curvature, respectively. Task letters and training-dataset letters identify different quantities.

The evaluation set $E_A$ is drawn from B after excluding all 312 episodes in A, leaving 738 eligible episodes. With random seed 42, we sample 13 episodes per task without replacement, yielding 78 episodes and 15,814 processed observation--action pairs. The sampling uses independent task-specific PCG64 generators initialized by \texttt{SeedSequence([42, task\_index])}, with tasks ordered T0, T1, T2, A, B and C and eligible source episode IDs sorted before selection. This split provides an evaluation set outside Ours-A's adaptation episodes, with equal episode counts across tasks. Table~\ref{tab:si-data} reports complete pair counts for $E_A$; Sec.~\ref{sec:si-offline} specifies frame sampling.

Datasets A and B use LeRobot v2.1 format, including synchronized RGB observations at $640\times480$, language instructions, 12-dimensional robot state and two continuous action components. Video runs at 20 fps. Temporal interpolation contributes to the processed-pair count: for task B, 28,590 captured frames correspond to 50,536 processed frames. These counts describe temporally related training entries.

The datasets store a 12-dimensional robot state and two action values per timestep. The state order is \texttt{rel\_x}, \texttt{rel\_y}, \texttt{rel\_z}, \texttt{abs\_x}, \texttt{abs\_y}, \texttt{abs\_z}, \texttt{qw}, \texttt{qx}, \texttt{qy}, \texttt{qz}, \texttt{m1\_pos}, \texttt{m2\_pos}. These fields encode relative position (3), absolute position (3), orientation (4) and bending-motor position proxies (2). Relative position is measured from the configured episode origin. Normalization uses training-data statistics; the same component order and transformation are applied online. All 312 episodes in A are assigned to training. $E_A$ provides episodes outside A for offline evaluation (Sec.~\ref{sec:si-offline}); physical evaluation uses separate robot trials.
\clearpage

\section{Robotic embodiments, policy interfaces and adaptation}\label{sec:si-robots}

\begin{center}\includegraphics[width=\linewidth,height=255pt,keepaspectratio]{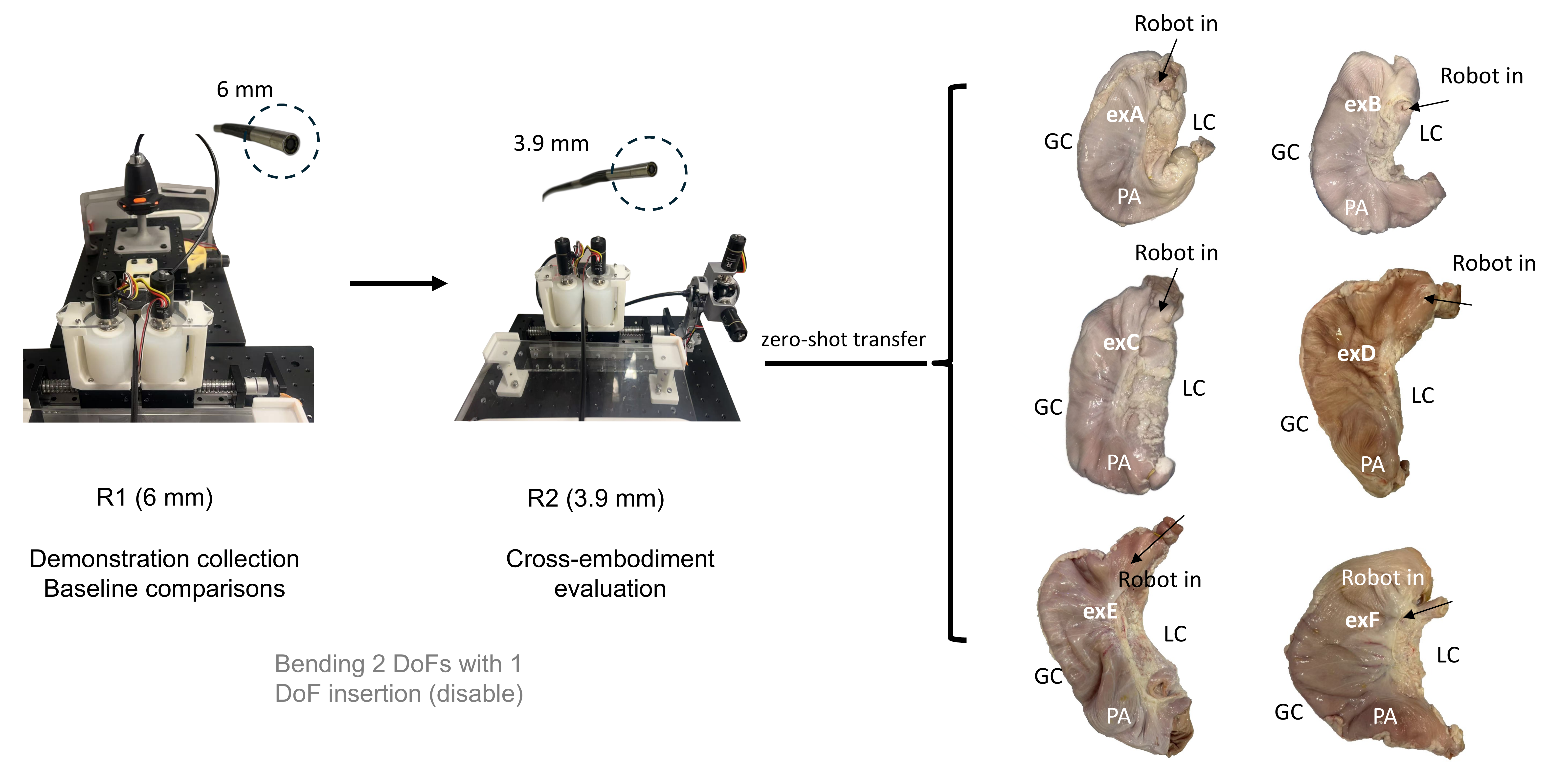}
\captionof{figure}{Robot platforms and ex-vivo evaluation settings. R1 (left) provides demonstrations and baseline/environment-transfer trials. R2 (middle) is used for the exA--exF experiments (right).}\label{fig:si-robots}
\end{center}

Both robots expose two distal bending commands, providing a common action interface across the 6 mm and 3.9 mm platforms. Insertion and shaft rotation remain fixed during autonomous model trials; the operator establishes the starting configuration before policy execution. Figure~\ref{fig:si-robots} connects the shared control interface to the change in physical embodiment and tissue environment evaluated in the main manuscript.

\setcounter{table}{1}
\renewcommand{\thetable}{S2}
\captionof{table}{Collection-software configuration for R1 and R2.}\label{tab:si-robots}
{\footnotesize\setlength{\tabcolsep}{4pt}\renewcommand{\arraystretch}{1.18}\begin{center}\begin{tabular}{p{\dimexpr 0.2500\linewidth-2\tabcolsep\relax}p{\dimexpr 0.3750\linewidth-2\tabcolsep\relax}p{\dimexpr 0.3750\linewidth-2\tabcolsep\relax}}\toprule\rowcolor{tableblue}

\color{white}\bfseries Item & \color{white}\bfseries R1 & \color{white}\bfseries R2 \\\midrule

Entry/config files & main.py; config.py & main.py; config.py \\

Camera frame size & 640 x 480 & 640 x 480 \\

Configured recording rate & 20 video fps; 20 data Hz & 20 video fps; 20 data Hz \\

Serial baud rate & 115200 & 115200 \\

Joystick bending mapping & m3 = horizontal x max\_speed; m4 = -vertical x max\_speed & Same mapping with M34\_SPEED\_SCALE = 0.3 \\

\bottomrule\end{tabular}\end{center}}

Table~\ref{tab:si-robots} documents the recording and joystick interfaces. Both programs record video and data at 20 Hz. R2 multiplies bending-velocity commands by 0.3 in both joystick control and autonomous policy deployment. For policy output, this factor is applied to the motor-velocity command after action denormalization. R2 responds more sensitively to motor-command changes, so the fixed scaling reduces bending speed. The policy-inference rates are measured separately in Sec.~\ref{sec:si-timing}.

The collection programs separate camera streaming, input handling, serial communication, recording and display. Online motor dispatch consumes one action every 50 ms, corresponding to 20 Hz.

\clearpage
\subsection{Policy inputs and adaptation}
The Ours-A/baseline evaluation is a \textbf{system-level comparison}. Each complete system uses dataset A and the same physical trial conditions, with its own observation representation, action formulation and adaptation procedure. This comparison measures the performance of the resulting endoscopic-control systems.

EndoVLA-OOV uses three temporally distinct images: the operator-established global observation at episode start, one recent-history frame randomly sampled from images acquired during the executed portion of the preceding action chunk, and the current decision image. The images share a visual encoder and enter the VLM with the instruction. A learned state encoder maps the normalized 12-dimensional continuous state to features supplied directly to the action expert. The expert conditions on these features, the visual--language context, noisy actions and flow time to predict a $20\times2$ bending-velocity chunk~\cite{lipman2023flow}. LoRA~\cite{hu2022lora} adapts the backbone, with trainable state and action interfaces.

The $\pi_0$ baseline~\cite{black2024pi_0} initializes from the official \texttt{pi0\_base} checkpoint and is fine-tuned on dataset A. It uses \texttt{Pi0Config} with \texttt{pi05=False}, \texttt{action\_dim=32} and \texttt{action\_horizon=20}, retaining the standard 32-dimensional model interface. Its visual input comprises the same initial/global, recent-history and current images as EndoVLA-OOV. The normalized 12-dimensional continuous state and two-dimensional training actions are independently zero-padded to width 32, adding 20 and 30 zeros, respectively. The padded state is linearly projected into a state token supplied to the action expert. The retained action head outputs 32 coordinates per step, of which the first two provide the physical bending commands. This configuration is used for both the success-rate evaluation and inference benchmark, with $H=20$ matching Ours.

The $\pi_{0.5}$ adaptation~\cite{pmlr-v305-black25a} starts from the official \texttt{pi05\_base} checkpoint and receives the same three temporal images. Normalized state values are discretized, serialized as numerical strings and processed by the language tokenizer, supplying state information through the VLM input. The standard OpenPI action interface has width 32 and supports padding a lower-dimensional action: $[0.2,-0.4]\mapsto[0.2,-0.4,0,\ldots,0]\in\mathbb{R}^{32}$. The retained 32-dimensional $\pi_{0.5}$ interface uses the first two output coordinates for endoscopic bending; $H=20$ matches Ours. Padding changes the representation width while preserving two physical commands.

For both $\pi_0$ and $\pi_{0.5}$, LoRA is applied to the attention and feed-forward layers of the PaliGemma language transformer and action-expert transformer, using ranks of 16 and 32, respectively. Their pretrained non-adapter parameters are frozen.

OpenVLA~\cite{pmlr-v270-kim25c} uses its published LoRA fine-tuning approach. Its input is the current image and instruction. Each of the two action values is discretized into an action token, retaining the language-model output head and generating two action tokens for one control step. Thus $H=1$. The EndoVLA comparison~\cite{pmlr-v305-kit25a} uses a Qwen2.5-VL backbone, the current image and instruction, and supervised fine-tuning followed by group relative policy optimization (GRPO). The reinforcement-learning reward sums L1-based action alignment and output-format accuracy with weight 1 for each component, giving a total reward in $[0,2]$. It generates one endoscopic action per policy query ($H=1$). These model-specific interfaces retain the corresponding continuous or token-based action formulation.
\clearpage

\subsection{Training data, optimization and checkpoint selection}
\label{sec:si-training}
\textbf{Common supervised data and budget.} All baselines and Ours-A use dataset A: six tasks with 52 episodes each, comprising 312 episodes and 79,310 observation anchors. Ours-B uses corpus B. Each model receives 40,000 supervised/flow-matching optimizer updates at effective batch 32. Every training job uses one NVIDIA GeForce RTX 5090 with 32 GB memory, a microbatch of four anchors and eight gradient-accumulation steps. Ours-A and Ours-B share the architecture, optimizer, update budget and checkpoint rule.

\textbf{Sampling and checkpoint selection.} Anchors are sampled uniformly with seed 42. Dataset-A comparisons use the same anchor-ordering rule across models. One anchor supplies a single action for token policies or a 20-action target for chunk policies. The supervised budget gives 1.28 million anchor exposures, equivalent to 16.14 passes over A or 4.89 over B. Temporally neighboring anchors and overlapping action chunks remain statistically dependent. Checkpoints are saved every 5,000 updates, and evaluation uses the final fixed-budget checkpoint. EndoVLA's final SFT checkpoint initializes its GRPO stage; evaluation uses the final GRPO checkpoint. Physical success rates are excluded from checkpoint selection.

\renewcommand{\thetable}{S2a}
\captionof{table}{Optimization settings and training budgets. EndoVLA's additional GRPO budget is reported separately from the common supervised budget.}
\label{tab:si-schedule}
{\small\setlength{\tabcolsep}{4pt}\renewcommand{\arraystretch}{1.2}
\begin{tabularx}{\linewidth}{@{}p{.15\linewidth}p{.24\linewidth}p{.18\linewidth}X@{}}
\toprule
Model & Learning rate and schedule & Effective batch & Optimizer updates \\
\midrule
$\pi_0$ & $2.5\!\times\!10^{-5}$ peak; cosine & 32 anchors & 40,000 \\
$\pi_{0.5}$ & $2.5\!\times\!10^{-5}$ peak; cosine & 32 anchors & 40,000 \\
Ours-A / B & $2.5\!\times\!10^{-5}$ peak; cosine & 32 anchors & 40,000 each \\
OpenVLA & $5\!\times\!10^{-4}$; constant & 32 anchors & 40,000 \\
EndoVLA SFT & $2\!\times\!10^{-4}$ peak; linear & 32 anchors & 40,000 \\
EndoVLA GRPO & $1\!\times\!10^{-5}$ peak; linear & 32 prompts; 4 responses each & 10,000 additional \\
\bottomrule
\end{tabularx}}

\textbf{Optimizer and numerics.} The OpenPI-derived policies use AdamW with $(\beta_1,\beta_2)=(0.9,0.95)$, $\epsilon=10^{-8}$ and weight decay $10^{-10}$. Their 1,000-update warmup reaches $2.5\times10^{-5}$, followed by cosine decay to $2.5\times10^{-6}$ at update 40,000. OpenVLA and EndoVLA use AdamW with $(0.9,0.999)$, $\epsilon=10^{-8}$ and weight decay 0.01. EndoVLA uses 3\% warmup followed by linear decay to zero within each stage. All jobs use BF16 original weights and forward computation, FP32 trainable parameters and optimizer moments, gradient checkpointing, gradient-norm clipping at 1.0, and disabled EMA. EndoVLA's GRPO stage generates 128 responses per update, adding a distinct generation and optimization budget to its supervised adaptation.

\textbf{Common action normalization.} Each action axis uses its 1st and 99th percentile statistics computed exclusively from dataset A:
\[
\overline a_j=2\frac{a_j-q_{01,j}}{q_{99,j}-q_{01,j}}-1,\qquad j\in\{1,2\}.
\]
The same transformation is used across the compared systems, including Ours-B. Token-policy targets are additionally clipped to their encoding range $[-1,1]$. The $\pi_0$/$\pi_{0.5}$ action interface pads the two normalized physical coordinates to width 32; flow loss selects only the two physical coordinates and uses the same per-action normalization as Ours. Inverse normalization recovers bending commands for motor dispatch.

\textbf{Implementation basis.} Learning-rate magnitudes follow the \href{https://github.com/Physical-Intelligence/openpi/blob/215abfb217dbac7d5f1273282331b9b1866c0479/src/openpi/training/optimizer.py}{OpenPI optimizer}, the \href{https://github.com/openvla/openvla/blob/c8f03f48af692657d3060c19588038c7220e9af9/vla-scripts/finetune.py}{OpenVLA fine-tuning script}, and the EndoVLA dual-phase recipe~\cite{pmlr-v305-kit25a}. Table~\ref{tab:si-schedule} reports the schedules used for the present endoscopic adaptations.
\clearpage

\subsection{Trainable and frozen parameters}
\label{sec:si-scope}
\renewcommand{\thetable}{S2b}
\captionof{table}{Parameter-update scope. ``Frozen'' refers to the original pretrained tensors; adapter tensors attached to a frozen layer remain trainable.}
\label{tab:si-scope}
{\small\setlength{\tabcolsep}{4pt}\renewcommand{\arraystretch}{1.18}
\begin{tabularx}{\linewidth}{@{}p{.13\linewidth}X X@{}}
\toprule
Model & Trainable parameters & Frozen parameters \\
\midrule
$\pi_0$ & LoRA factors on attention projections and feed-forward matrices in all PaliGemma language-transformer layers ($r=16$, $\alpha=16$) and action-expert layers ($r=32$, $\alpha=32$). & All original weights: visual encoder, image projection, token embeddings, both transformers and their normalization parameters, continuous-state projection, action input/output projections and action/time MLPs. \\
\addlinespace
$\pi_{0.5}$ & The same attention/feed-forward LoRA scopes and ranks as $\pi_0$. & All original weights: visual encoder and image projection, embeddings, both transformers, normalization and adaptive-normalization parameters, time-conditioning MLPs, and action input/output projections. \\
\addlinespace
Ours-A / B & Language-transformer attention/feed-forward LoRA ($r=16$, $\alpha=16$); action-expert attention/feed-forward LoRA ($r=32$, $\alpha=32$); all weights and biases of the continuous-state MLP, 2D action-input projection and 2D action-output MLP. & Original visual encoder and image projection, token embeddings, language/expert transformer tensors, normalizations, time-conditioning MLPs and adaptive-normalization parameters. The replaced action projections are excluded from the active model. \\
\addlinespace
OpenVLA & LoRA ($r=32$, $\alpha=16$, dropout 0) on the linear modules selected by PEFT's \texttt{all-linear} rule: language and vision attention/MLP linears and multimodal-projector linears. & Original weights of the adapted layers, embeddings, convolutional components, normalization parameters and the language-model output head. The existing action-token vocabulary is retained. \\
\addlinespace
EndoVLA & $r=32$, $\alpha=32$, dropout 0 LoRA on Qwen2.5-VL language attention (query/key/value/output) and feed-forward (gate/up/down) projections; all weights and biases of the visual-to-language merger. The same scope is updated in SFT and GRPO. & Visual feature extractor outside the merger; original language weights, embeddings, normalizations and output head. GRPO uses a frozen copy of the SFT policy as its reference. \\
\bottomrule
\end{tabularx}}

\textbf{Implementation of the freeze policy.} For $\pi_0$ and $\pi_{0.5}$, an explicit trainable-parameter allowlist freezes all pretrained non-adapter tensors. The \href{https://github.com/Physical-Intelligence/openpi/blob/215abfb217dbac7d5f1273282331b9b1866c0479/src/openpi/models/pi0_config.py}{OpenPI LoRA freeze filter} selects transformer paths; the visual and interface modules outside those paths are additionally frozen to obtain the stated scope. Ours extends that allowlist with the newly introduced state and action modules. The OpenPI \href{https://github.com/Physical-Intelligence/openpi/blob/215abfb217dbac7d5f1273282331b9b1866c0479/src/openpi/models/gemma.py}{Gemma configurations} define the ranks and scaling factors.

\textbf{Model-specific details.} OpenVLA follows the official all-linear PEFT configuration, including eligible vision and projector layers; its output head keeps the existing token mapping. EndoVLA initializes from Qwen2.5-VL-7B-Instruct. Its published recipe motivates language LoRA and a trainable visual-to-language projector~\cite{pmlr-v305-kit25a}; the Qwen2.5-VL merger and ranks above are the implementation choices. The GRPO reward is $R=R_{\rm align}+R_{\rm format}\in[0,2]$: L1-based action alignment and format accuracy each have coefficient 1. The parameter scopes in Table~\ref{tab:si-scope} define the tensors updated in each adaptation.
\clearpage

\subsection{Ours state/action connection}
\label{sec:si-connection}
\textbf{Visual and language prefix.} The model processes the initial global, recent-history and current images through the shared pretrained visual encoder in that fixed order. It concatenates their visual tokens with the instruction tokens as the VLM prefix. The pretrained image preprocessing and positional-index convention are retained. An instruction-only token stream supplies language; the normalized continuous state enters through the expert pathway below. The $\pi_{0.5}$-based implementation uses an instruction-only prompt and retains flow-time conditioning. The pretrained language and expert hidden widths are 2,048 and 1,024, respectively, as specified by the OpenPI Gemma configurations.

\textbf{Independent state and action widths.} The interface declares separate widths $d_s=12$ and $d_a=2$. This decouples the state encoder from the action-input width. The state encoder maps the normalized state to one expert-width token using two affine layers and a SiLU activation:
\[
\mathbf{z}_s=W_{s2}\,\operatorname{SiLU}(W_{s1}\overline{\mathbf{s}}_t+\mathbf{b}_{s1})+\mathbf{b}_{s2},
\qquad 12\longrightarrow256\longrightarrow1024.
\]
Each noisy two-dimensional action is projected into the same expert width with a trainable affine map $2\rightarrow1024$. The resulting suffix has one state token followed by 20 action tokens. A separate state-width specification replaces the standard shared state/action shape assumption in the model input declaration and preprocessing.

\textbf{Attention and time conditioning.} The prefix attends within its image--language context. The state token attends to the prefix and itself. The action tokens attend to the prefix, the state token and all action tokens in the same predicted chunk. This block mask keeps the prefix independent of the suffix, supporting prefix key/value caching across flow-integration steps. The action expert retains the pretrained $\pi_{0.5}$ time MLP and adaptive RMS-normalization pathway, with flow time supplying the expert's time condition. The shared attention head geometry connects the language and expert streams despite their different hidden widths.

\textbf{Two-dimensional flow output.} A trainable $1024\rightarrow256\rightarrow2$ MLP, with SiLU between its affine layers, maps each of the 20 action-token representations to two flow coordinates. The model selects these final 20 token positions for the flow loss and predicted velocity field. The state-token output serves as context and is excluded from action loss. Flow integration operates on a $20\times2$ noisy-action tensor; inverse action normalization maps the resulting two coordinates to the bending interface. The model input/output declarations, action normalization and flow loss all retain this two-dimensional action shape.

\textbf{Initialization and parameter updates.} The model loads compatible pretrained visual, language, expert and time-conditioning tensors from \texttt{pi05\_base}. It introduces the state MLP, two-dimensional action-input projection and two-dimensional output MLP as new modules, using Xavier-uniform weight initialization and zero biases. The original high-dimensional action projections are replaced. The adaptation applies language/expert LoRA with ranks 16/32 and updates the new interface modules; the remaining original tensors stay frozen according to Table~\ref{tab:si-scope}. Ours-A and Ours-B use the same connectivity and update scope.

\textbf{Connection to deployment.} The execution rule uses back-to-back inference and a 20 Hz motor-dispatch loop. A completed prediction replaces the unexecuted remainder of the previous chunk. When the sequence is exhausted, the last velocity command is held until the next prediction arrives. The recent-history image is sampled from images collected during that preceding chunk's executed portion. Thus the expert predicts a 20-step horizon while fresh observations update the active sequence as subsequent inference results arrive. The sequence of accepted predictions determines how many commands are executed from each chunk.

The reference \href{https://github.com/Physical-Intelligence/openpi/blob/215abfb217dbac7d5f1273282331b9b1866c0479/src/openpi/models/pi0.py}{OpenPI model implementation} provides the action-expert, prefix/suffix attention and flow-time components. The continuous-state MLP, independent state/action widths and two-dimensional interfaces above specify the endoscopic adaptation.
\clearpage

\section{Detector-assisted annotation and human review}\label{sec:si-annotation}
The evaluation uses an external YOLOv8n detector to propose target bounding boxes. YOLO Studio supports manual box creation, detector fine-tuning, automatic annotation and frame-by-frame review. After inference, all frame records and proposed boxes were manually reviewed and corrected as needed. Policy actions use the image, instruction and state interface described in the main manuscript; the detector supports evaluation annotation.

\begin{center}\includegraphics[width=\linewidth,height=165pt,keepaspectratio]{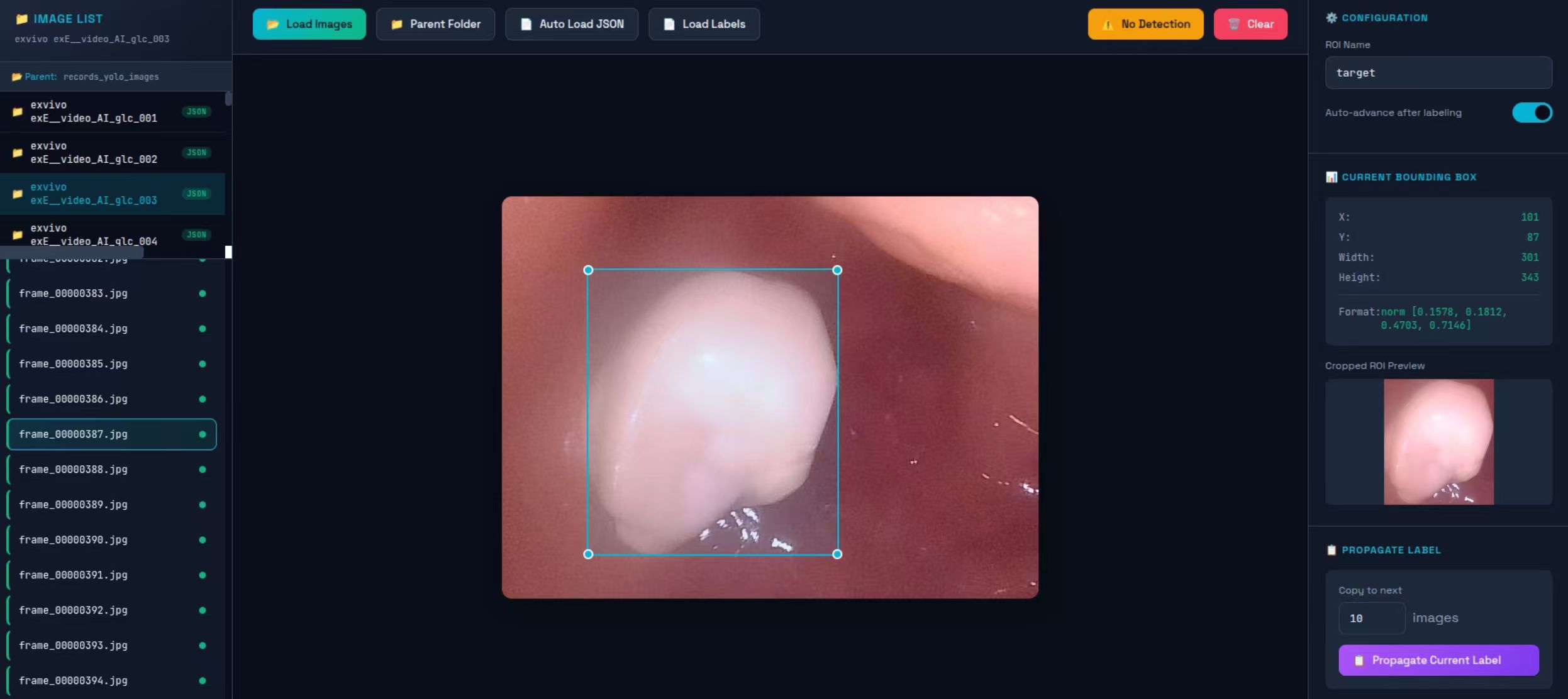}
\captionof{figure}{YOLO Studio annotation interface. Editable target boxes and an explicit No Detection control support frame-level verification of target visibility and position.}\label{fig:si-ui}
\end{center}

\setcounter{table}{2}
\renewcommand{\thetable}{S3}
\captionof{table}{Internal detector-development annotation corpora.}\label{tab:si-annotation}
{\footnotesize\setlength{\tabcolsep}{4pt}\renewcommand{\arraystretch}{1.18}\begin{center}\begin{tabular}{p{\dimexpr 0.2900\linewidth-2\tabcolsep\relax}p{\dimexpr 0.1400\linewidth-2\tabcolsep\relax}p{\dimexpr 0.1900\linewidth-2\tabcolsep\relax}p{\dimexpr 0.1900\linewidth-2\tabcolsep\relax}p{\dimexpr 0.1900\linewidth-2\tabcolsep\relax}}\toprule\rowcolor{tableblue}

\color{white}\bfseries Internal detector corpus & \color{white}\bfseries Episodes & \color{white}\bfseries Frames & \color{white}\bfseries Box frames & \color{white}\bfseries no\_detection \\\midrule

Clear lesions & 117 & 74,163 & 73,235 & 928 \\

Mucus occlusion & 427 & 65,673 & 65,605 & 68 \\

Total & 544 & 139,836 & 138,840 & 996 \\

\bottomrule\end{tabular}\end{center}}

\setcounter{table}{3}
\renewcommand{\thetable}{S4}
\captionof{table}{YOLOv8n training and internal validation results.}\label{tab:si-detector}
{\footnotesize\setlength{\tabcolsep}{4pt}\renewcommand{\arraystretch}{1.18}\begin{center}\begin{tabular}{p{\dimexpr 0.2400\linewidth-2\tabcolsep\relax}p{\dimexpr 0.1700\linewidth-2\tabcolsep\relax}p{\dimexpr 0.1700\linewidth-2\tabcolsep\relax}p{\dimexpr 0.2100\linewidth-2\tabcolsep\relax}p{\dimexpr 0.2100\linewidth-2\tabcolsep\relax}}\toprule\rowcolor{tableblue}

\color{white}\bfseries YOLOv8n run & \color{white}\bfseries Train entries & \color{white}\bfseries Val entries & \color{white}\bfseries Best mAP50 & \color{white}\bfseries Best mAP50-95 \\\midrule

Mucus occlusion & 12,222 & 1,357 & 0.99495 & 0.93641 \\

Clear lesions & 3,393 & 376 & 0.99500 & 0.95727 \\

\bottomrule\end{tabular}\end{center}}

Table~\ref{tab:si-annotation} accounts for all 139,836 frame records across 544 episodes: 138,840 contain a box and 996 carry a \texttt{no\_detection} entry. Recording both outcomes makes target visibility explicit during scoring. Figure~\ref{fig:si-ui} shows the interface used to inspect the corresponding frame and box records.

Both detector runs used 50 epochs and batch size 16. Table~\ref{tab:si-detector} lists selected training/validation image entries and the best logged values for each metric. mAP50 is approximately 0.995 for both runs, while mAP50--95 is 0.93641 for mucus occlusion and 0.95727 for clear lesions. These internal validation results support the use of the detector to generate annotation candidates; human review supplies the frame-level verification for the robot evaluations.
\clearpage

\section{Instructions and evaluation protocols}\label{sec:si-protocols}

\setcounter{table}{4}
\renewcommand{\thetable}{S5}
\captionof{table}{Standard ex-vivo instructions, reproduced verbatim.}\label{tab:si-prompts}
{\footnotesize\setlength{\tabcolsep}{4pt}\renewcommand{\arraystretch}{1.18}\begin{center}\begin{tabular}{p{\dimexpr 0.1800\linewidth-2\tabcolsep\relax}p{\dimexpr 0.2300\linewidth-2\tabcolsep\relax}p{\dimexpr 0.5900\linewidth-2\tabcolsep\relax}}\toprule\rowcolor{tableblue}

\color{white}\bfseries Organs & \color{white}\bfseries Region & \color{white}\bfseries Instruction \\\midrule

exA / exD & Greater Curvature & Search the Greater Curvature for a orange oval suspicious region on the mucosal folds. \\

exB / exE & Pyloric Antrum & Search the Pyloric Antrum to find a small white round raised nodule. \\

exC / exF & Lesser Curvature & Search the Lesser Curvature and locate the Orange-white protruding lesion. \\

\bottomrule\end{tabular}\end{center}}

Table~\ref{tab:si-prompts} pairs exA/exD, exB/exE and exC/exF through the same anatomical-region and appearance request. This pairing specifies the language objective for both policies in the exD--exF corpus comparison. The strings preserve the wording used for evaluation, including their capitalization and grammar.

\subsection{Instruction and occlusion conditions}
The full-instruction condition on exA uses its standard prompt. The unseen-correct-wording condition on exB alternates the following two prompts, beginning with the first:
\begin{quote}\small
Inspect the Pyloric Antrum for a pale, Circular mucosal abnormality seated on the folds.\\[3pt]
Along the Pyloric Antrum, localize the whitish Circular region of altered mucosa.
\end{quote}
The empty-instruction condition on exC sends an empty string. The wrong-instruction condition on exA alternates the following requests, beginning with the first:
\begin{quote}\small
Search the Pyloric Antrum to find a small round raised nodule.\\[3pt]
Search the Lesser Curvature and locate the orange protruding lesion.
\end{quote}
The wrong-instruction intervention changes both region and target appearance; scoring uses the reference target specified by the standard exA prompt. Results are reported for each instruction condition. Post-occlusion recovery on exC retains its standard prompt and evaluates reacquisition after visibility is interrupted.

\subsection{Find and Center endpoints}
R1, R2 and the training-phantom input interventions share the same Find criterion. Within a trial horizon of 3,000 frames (150 seconds at 20 fps), the reference target must be continuously visible in at least 100 consecutive human-reviewed frames, corresponding to five seconds. $SR_{\rm find}$ is the number of successful trials divided by all attempted trials. After Find is satisfied, composite finding-and-tracking, $SR_{\rm F\&T}$, requires the target-box center to enter the central rectangle of width $0.5W$ and height $0.5H_{\rm img}$, with target visibility retained through the remaining trial. Center success also requires prior Find success. It then requires the target-box center $(x_c,y_c)$ to remain in that same rectangle for two continuous seconds with uninterrupted visibility:
\begin{equation}
0.25\le x_c/W\le0.75,\qquad 0.25\le y_c/H_{\rm img}\le0.75.
\label{eq:si-center}
\end{equation}
Here $W$ and $H_{\rm img}$ are image dimensions; the rectangle occupies 25\% of image area. Each endpoint uses all attempted trials as the denominator. Standard exA and the separate full-instruction exA group each have Center 8/10. The instruction conditions use the organ assignments shown in the main-text figure; comparisons are interpreted within those assignments.

\subsection{Training-phantom input interventions}
All four view conditions use the same Ours-B checkpoint and the prompt ``Search the Greater Curvature for a white oval suspicious region on the mucosal folds.'' We retain the current image (C) and mask neither, one or both contextual slots: G+R+C, R+C (global masked), G+C (recent masked), and C only (both masked). G and R denote global and recent-history images. Each condition contains ten trials; (Find, Center) success counts are respectively (9, 9), (7, 6), (7, 7) and (6, 5), each out of ten (main Table~III). The current-only condition supplies one informative image while preserving the three-slot interface. Center follows Eq.~\eqref{eq:si-center}. The reciprocal competing-target requests select the GC white oval while ignoring the PA small round target, or vice versa; each achieves Find 7/10.
\clearpage

\section{Offline instruction and action analyses}\label{sec:si-offline}
\setlength{\parskip}{3pt}
The offline analyses evaluate Ours-A, trained on dataset A, on the episode-disjoint set $E_A$: 78 episodes and 15,814 processed pairs (Sec.~\ref{sec:si-data}). Sampling restarts at frame 0 in each episode and selects indices $0,8,16,\ldots$. With $L_e$ denoting the processed-pair count of episode $e$,
\[
N_{\mathrm{sampled}}=\sum_{e=1}^{78}\left\lceil\frac{L_e}{8}\right\rceil=2{,}012.
\]
The counts for T0, T1, T2, A, B and C are 317, 300, 368, 217, 345 and 465. Each sampled frame is evaluated under nine instruction conditions, yielding 18,108 predicted action chunks. Frames within an episode remain temporally related.

\subsection{Substitution of trained instructions}
We compare the predicted $20\times2$ action chunk under the reference and substituted instructions. A single Gaussian noise tensor from \texttt{numpy.random.default\_rng(0)} is reused across all frames and conditions. At each frame, visual observations, the 12-dimensional state, checkpoint and noise are fixed; only the instruction changes. The mean chunk distance is
\begin{equation}
D_{ij}=\frac{1}{N_i}\sum_{n=1}^{N_i}
\left\|\widehat{\mathbf A}_{n}^{(i)}-\widehat{\mathbf A}_{n}^{(j)}\right\|_{F},
\label{eq:si-l2}
\end{equation}
where $i$ identifies the recorded task, $j$ the replacement instruction and $N_i$ the sampled-frame count. The Frobenius norm is the L2 norm of the 40 action components.

\begin{center}\includegraphics[width=\linewidth,height=190pt,keepaspectratio]{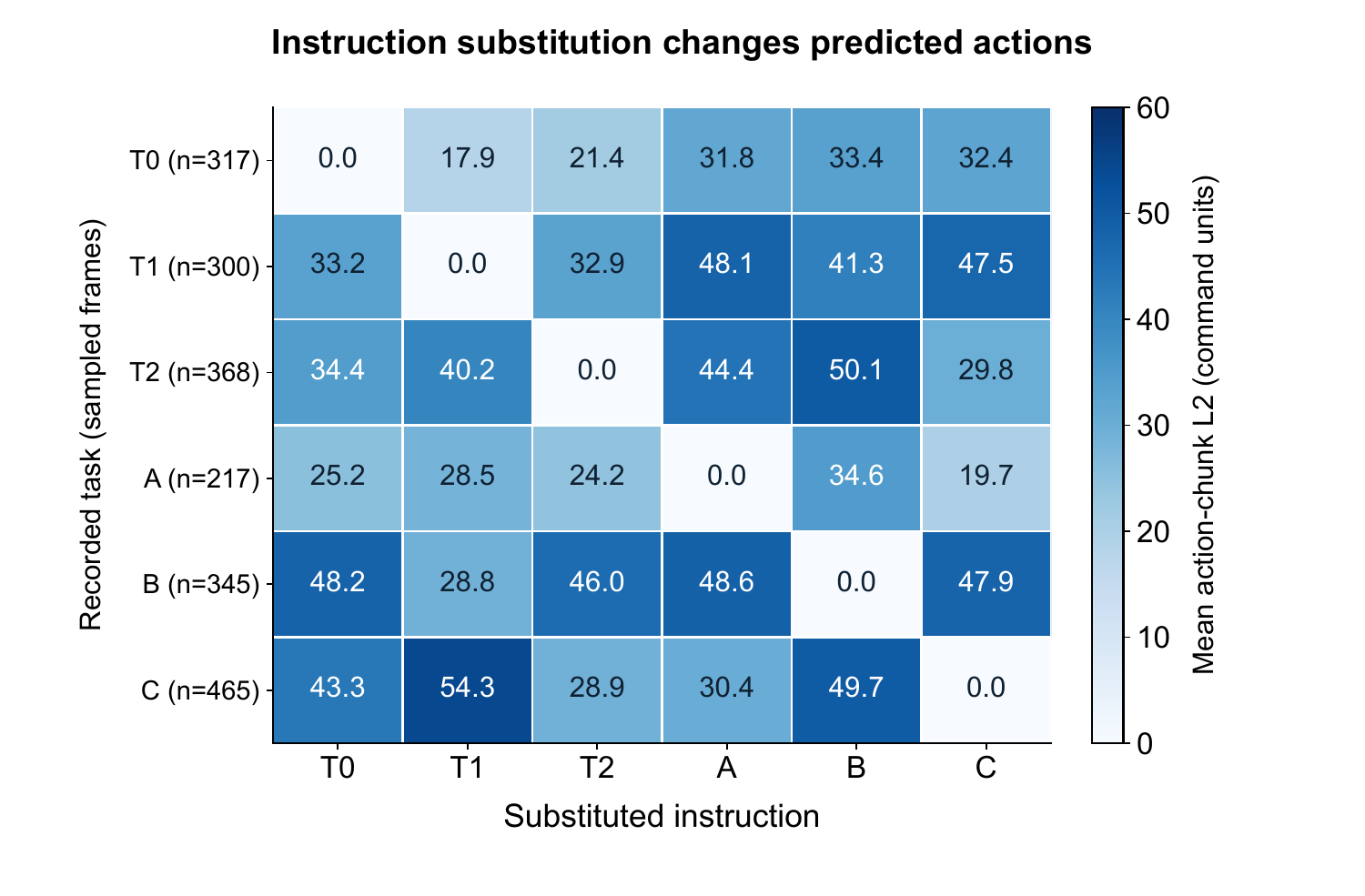}
\captionof{figure}{Ours-A action changes under trained-instruction substitution on $E_A$. Each cell is the mean chunk distance in Eq.~\eqref{eq:si-l2}. Rows identify recorded tasks and columns identify replacement instructions. The diagonal is the unchanged-instruction reference.}\label{fig:si-swap}
\end{center}

\setcounter{table}{5}\renewcommand{\thetable}{S6a}
\captionof{table}{Mean action-chunk L2 distance after instruction substitution.}\label{tab:si-swap}
{\footnotesize\setlength{\tabcolsep}{4pt}\renewcommand{\arraystretch}{1.18}\begin{center}\begin{tabular}{p{\dimexpr 0.0900\linewidth-2\tabcolsep\relax}p{\dimexpr 0.1300\linewidth-2\tabcolsep\relax}p{\dimexpr 0.1300\linewidth-2\tabcolsep\relax}p{\dimexpr 0.1300\linewidth-2\tabcolsep\relax}p{\dimexpr 0.1300\linewidth-2\tabcolsep\relax}p{\dimexpr 0.1300\linewidth-2\tabcolsep\relax}p{\dimexpr 0.1300\linewidth-2\tabcolsep\relax}p{\dimexpr 0.1300\linewidth-2\tabcolsep\relax}}\toprule\rowcolor{tableblue}
\color{white}\bfseries Task & \color{white}\bfseries Frames & \color{white}\bfseries T0 & \color{white}\bfseries T1 & \color{white}\bfseries T2 & \color{white}\bfseries A & \color{white}\bfseries B & \color{white}\bfseries C \\\midrule
T0 & 317 & 0.0 & 17.9 & 21.4 & 31.8 & 33.4 & 32.4 \\
T1 & 300 & 33.2 & 0.0 & 32.9 & 48.1 & 41.3 & 47.5 \\
T2 & 368 & 34.4 & 40.2 & 0.0 & 44.4 & 50.1 & 29.8 \\
A & 217 & 25.2 & 28.5 & 24.2 & 0.0 & 34.6 & 19.7 \\
B & 345 & 48.2 & 28.8 & 46.0 & 48.6 & 0.0 & 47.9 \\
C & 465 & 43.3 & 54.3 & 28.9 & 30.4 & 49.7 & 0.0 \\
\bottomrule\end{tabular}\end{center}}

Figure~\ref{fig:si-swap} and Table~\ref{tab:si-swap} show off-diagonal distances of 17.9--54.3 command units. C-to-T1 gives the largest change (54.3), followed by T2-to-B (50.1). These changes quantify instruction sensitivity with the other inference inputs fixed. Dividing each chunk distance by $\sqrt{40}$ gives its coordinate RMS; the corresponding mean values span 2.8--8.6 command units.
\clearpage

\setlength{\parskip}{5pt}
\subsection{Empty, shuffled and fixed unseen instructions}
Using the same fixed-noise protocol, we compare each reference prediction with an empty string, the fixed shuffled string ``region suspicious oval white Tracking.'', and the fixed unseen sentence ``Please look for the small bump near the antrum.'' The same strings are used across all tasks. The unseen sentence preserves the requested target for task B and changes the target for the other tasks. The exB correct-wording physical trials use the separate prompts in Sec.~\ref{sec:si-protocols}.

\begin{center}\includegraphics[width=\linewidth,height=215pt,keepaspectratio]{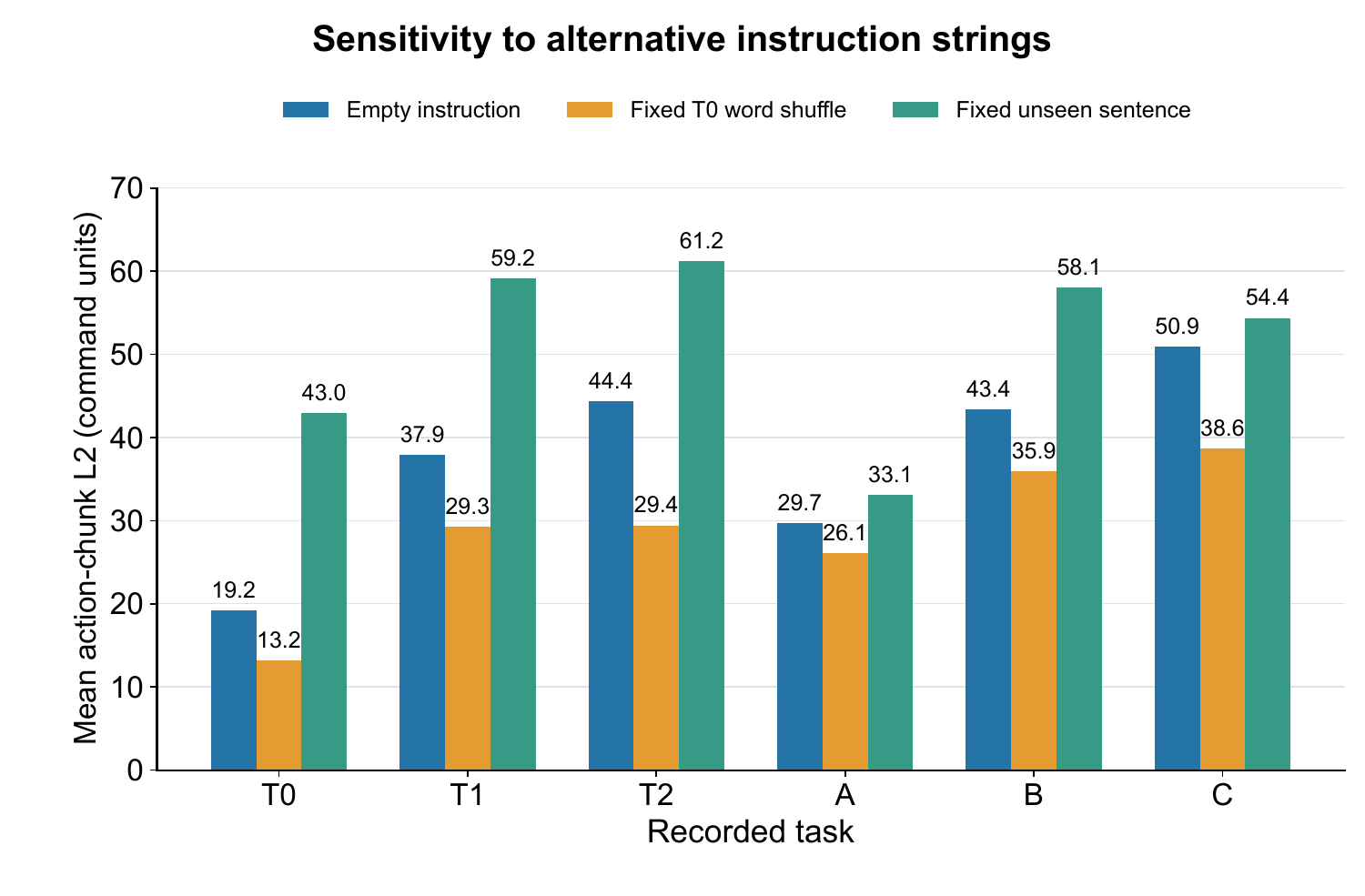}
\captionof{figure}{Ours-A action sensitivity to alternative instruction strings on $E_A$. Each bar is the mean action-chunk L2 distance from the reference prediction under one shared noise realization.}\label{fig:si-alt}
\end{center}

\setcounter{table}{5}\renewcommand{\thetable}{S6b}
\captionof{table}{Mean action-chunk L2 distance under alternative instruction strings.}\label{tab:si-alt}
{\footnotesize\setlength{\tabcolsep}{4pt}\renewcommand{\arraystretch}{1.18}\begin{center}\begin{tabular}{p{\dimexpr 0.1600\linewidth-2\tabcolsep\relax}p{\dimexpr 0.2000\linewidth-2\tabcolsep\relax}p{\dimexpr 0.3000\linewidth-2\tabcolsep\relax}p{\dimexpr 0.3400\linewidth-2\tabcolsep\relax}}\toprule\rowcolor{tableblue}
\color{white}\bfseries Task & \color{white}\bfseries Empty & \color{white}\bfseries Fixed T0 shuffle & \color{white}\bfseries Fixed unseen sentence \\\midrule
T0 & 19.2 & 13.2 & 43.0 \\
T1 & 37.9 & 29.3 & 59.2 \\
T2 & 44.4 & 29.4 & 61.2 \\
A & 29.7 & 26.1 & 33.1 \\
B & 43.4 & 35.9 & 58.1 \\
C & 50.9 & 38.6 & 54.4 \\
\bottomrule\end{tabular}\end{center}}

The fixed unseen sentence gives the largest distance for every task in Figure~\ref{fig:si-alt} and Table~\ref{tab:si-alt}, ranging from 33.1 to 61.2. Empty instructions give 19.2--50.9, and the fixed shuffle gives 13.2--38.6. T2 has the largest response to the unseen sentence (61.2), compared with 44.4 for the empty string and 29.4 for the shuffle.

The ordering shows that the content of the substituted string influences the predicted action change. These paired comparisons characterize the evaluated policy's use of language under fixed visual and state inputs. The physical instruction and competing-target trials provide the complementary evidence for acquiring the requested target. All offline estimates are conditional on one shared noise realization and the sampled episodes.
\clearpage

\subsection{Motor-direction agreement across chunk prefixes}
At each sampled anchor, direction agreement is the fraction of eligible motor components whose predicted and demonstrated signs match. Eligibility requires demonstrated absolute speed greater than 1 in command units. We average these fractions across anchors with at least one eligible component, separately for each task and prefix. Table~\ref{tab:si-direction} gives these eligible-anchor counts as $n$.

The prefixes include the first one, first two and up to twenty predicted actions. At episode ends, comparison uses the available recorded future actions. Of 2,012 anchors, 1,828 have all twenty targets and 184 have a shorter future horizon. This protocol evaluates predictions against recorded actions along the sequence.

\begin{center}\includegraphics[width=\linewidth,height=210pt,keepaspectratio]{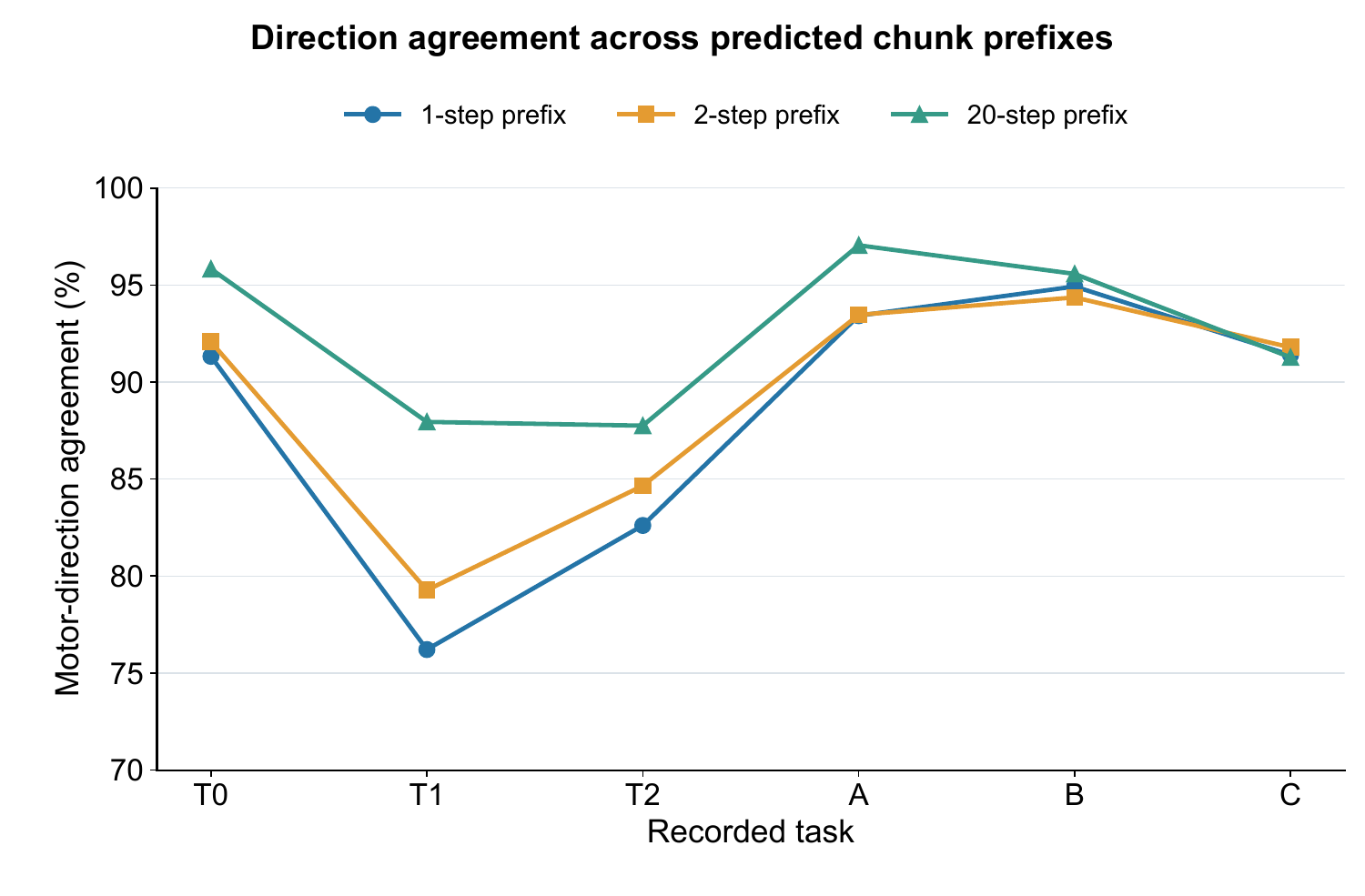}
\captionof{figure}{Motor-direction agreement over 1-, 2- and 20-step prediction prefixes. Each rate averages the per-anchor sign-agreement fraction on eligible demonstrated motor components.}\label{fig:si-direction}
\end{center}

\setcounter{table}{5}\renewcommand{\thetable}{S6c}
\captionof{table}{Direction agreement (\%) and eligible-anchor counts $n$ for each prefix.}\label{tab:si-direction}
{\footnotesize\setlength{\tabcolsep}{4pt}\renewcommand{\arraystretch}{1.18}\begin{center}\begin{tabular}{p{\dimexpr 0.1600\linewidth-2\tabcolsep\relax}p{\dimexpr 0.2800\linewidth-2\tabcolsep\relax}p{\dimexpr 0.2800\linewidth-2\tabcolsep\relax}p{\dimexpr 0.2800\linewidth-2\tabcolsep\relax}}\toprule\rowcolor{tableblue}
\color{white}\bfseries Task & \color{white}\bfseries 1-step prefix & \color{white}\bfseries 2-step prefix & \color{white}\bfseries 20-step prefix \\\midrule
T0 & 91.3\% ($n=219$) & 92.1\% ($n=227$) & 95.8\% ($n=280$) \\
T1 & 76.2\% ($n=227$) & 79.3\% ($n=234$) & 87.9\% ($n=277$) \\
T2 & 82.6\% ($n=276$) & 84.7\% ($n=290$) & 87.8\% ($n=352$) \\
A & 93.4\% ($n=152$) & 93.5\% ($n=153$) & 97.0\% ($n=184$) \\
B & 94.9\% ($n=266$) & 94.4\% ($n=269$) & 95.6\% ($n=309$) \\
C & 91.4\% ($n=337$) & 91.8\% ($n=341$) & 91.3\% ($n=407$) \\
\bottomrule\end{tabular}\end{center}}

Figure~\ref{fig:si-direction} and Table~\ref{tab:si-direction} show first-action agreement of 76.2--94.9\%, two-action agreement of 79.3--94.4\%, and full-prefix agreement of 87.8--97.0\%. T1 increases from 76.2\% to 87.9\%, and T2 from 82.6\% to 87.8\%. Tasks A and B reach 97.0\% and 95.6\% over the full prefix.

The rates show directional alignment with the recorded demonstrations, with task-dependent variation along the predicted sequence. Prefix lengths include different future coordinates and eligible anchors. These measurements characterize prediction agreement; the physical trials assess closed-loop target acquisition and centering.
\clearpage

\section{Perception rate and maximum chunk-equivalent control rate}\label{sec:si-timing}
Each of the five models is measured over 300 input-to-action-output inferences using common timing boundaries on an NVIDIA GeForce RTX 5090 GPU with 32 GB VRAM and an Intel Core Ultra 9 285K CPU. With mean inference duration $T_{\rm infer}$, total duration $T_{\rm total}$ and output chunk size $H$, we define
\begin{equation}
A=\frac{1}{T_{\rm infer}}=\frac{300}{T_{\rm total}},\qquad
B=\frac{H}{T_{\rm infer}}=AH.
\label{eq:si-rates}
\end{equation}
$A$ counts fresh inputs processed per second in the inference benchmark. $B$ is the theoretical maximum control rate when every action in a chunk is used between two successive processed observations separated by $T_{\rm infer}$. Equivalently, the $H$ actions are assumed to be evenly dispatched with interval $T_{\rm infer}/H=1/B$. This calculation describes the capacity of the generated sequence under that assumption. Every action in the chunk is conditioned on the same input; a further policy query incorporates a fresh observation.

\setcounter{table}{6}
\renewcommand{\thetable}{S7}
\captionof{table}{300-input inference benchmark and theoretical maximum chunk-equivalent control rate.}\label{tab:si-rates}
{\footnotesize\setlength{\tabcolsep}{4pt}\renewcommand{\arraystretch}{1.18}\begin{center}\begin{tabular}{p{\dimexpr 0.2300\linewidth-2\tabcolsep\relax}p{\dimexpr 0.1500\linewidth-2\tabcolsep\relax}p{\dimexpr 0.1000\linewidth-2\tabcolsep\relax}p{\dimexpr 0.2900\linewidth-2\tabcolsep\relax}p{\dimexpr 0.2300\linewidth-2\tabcolsep\relax}}\toprule\rowcolor{tableblue}

\color{white}\bfseries Model & \color{white}\bfseries A (Hz) & \color{white}\bfseries H & \color{white}\bfseries B = A x H (actions/s) & \color{white}\bfseries 1/A (ms) \\\midrule

EndoVLA & 1.480 & 1 & 1.480 & 675.7 \\

OpenVLA-7B & 5.228 & 1 & 5.228 & 191.3 \\

$\pi_0$ & 17.69 & 20 & 353.8 & 56.5 \\

$\pi_{0.5}$ & 15.68 & 20 & 313.6 & 63.8 \\

Ours & 14.27 & 20 & 285.4 & 70.1 \\

\bottomrule\end{tabular}\end{center}}

Table~\ref{tab:si-rates} reports perception rates of 17.69, 15.68 and 14.27 Hz for $\pi_0$, $\pi_{0.5}$ and Ours. With $H=20$, the corresponding theoretical maximum rates are 353.8, 313.6 and 285.4 actions/s. Ours has a mean inference latency of 70.1 ms; its idealized action interval under the definition of $B$ is 3.50 ms. EndoVLA and OpenVLA generate one action per input, giving $A=B=1.480$ and 5.228 Hz, respectively.

The physical endoscope dispatch interval is 50 ms, giving 20 motor commands/s, and the 20-action prediction horizon covers one second. Deployment uses back-to-back policy queries: each prediction is immediately followed by the next query with the latest available observation. Each completed chunk prediction replaces the unexecuted remainder of the preceding chunk. The motor-dispatch loop continues at 20 Hz alongside successive inferences. When the available sequence is exhausted, the most recent velocity is held until the next prediction arrives. For the $H=1$ EndoVLA and OpenVLA baselines, this holds the single predicted velocity throughout the interval between predictions. Accordingly, $H=20$ specifies the prediction horizon, and the number of actions actually executed from a chunk depends on the arrival time of the next prediction. Measured inference capacity, the idealized rate $B$ and physical dispatch remain separately specified quantities.

Camera capture, preprocessing, communication, queueing and actuator response contribute to end-to-end delay. The 300-input benchmark measures the input-to-output interval. Tracking accuracy additionally depends on frequent incorporation of fresh images and the age of feedback at action execution. ACT's temporal ensembling~\cite{Zhao-RSS-23} and real-time chunking~\cite{black2026real} coordinate successive predictions with execution. These mechanisms motivate evaluation of a future handover from semantic target acquisition to a controller with low perception latency, frequent fresh-observation updates and precise distal control.
\clearpage

\section{Supplementary camera-control experiments}\label{sec:si-control}
\subsection{Centering under lighting and wetness changes}
The R1 centering experiment begins with the target visible and runs for up to 300 frames (15 seconds at 20 fps). Ours-A and all VLA baselines are fine-tuned on subset A. Pixel error $PE_t$ is the Euclidean distance between the human-reviewed target-box center and image center, expressed in the $640\times480$ recorded-image coordinates. Frames labeled \texttt{no\_detection} receive the image diagonal of 800 pixels.

The plotted index summarizes the mean successive change in pixel error over a trial of $N$ frames:
\begin{equation}
\Delta PE=\frac{1}{N-1}\sum_{t=2}^{N}(PE_t-PE_{t-1})
=\frac{PE_N-PE_1}{N-1}.
\label{eq:si-deltape}
\end{equation}
Negative values indicate net reduction in pixel error; positive values indicate net increase. The index summarizes endpoint change, while the time-resolved traces describe intermediate corrections and visibility changes.

\begin{center}\includegraphics[width=\linewidth,height=250pt,keepaspectratio]{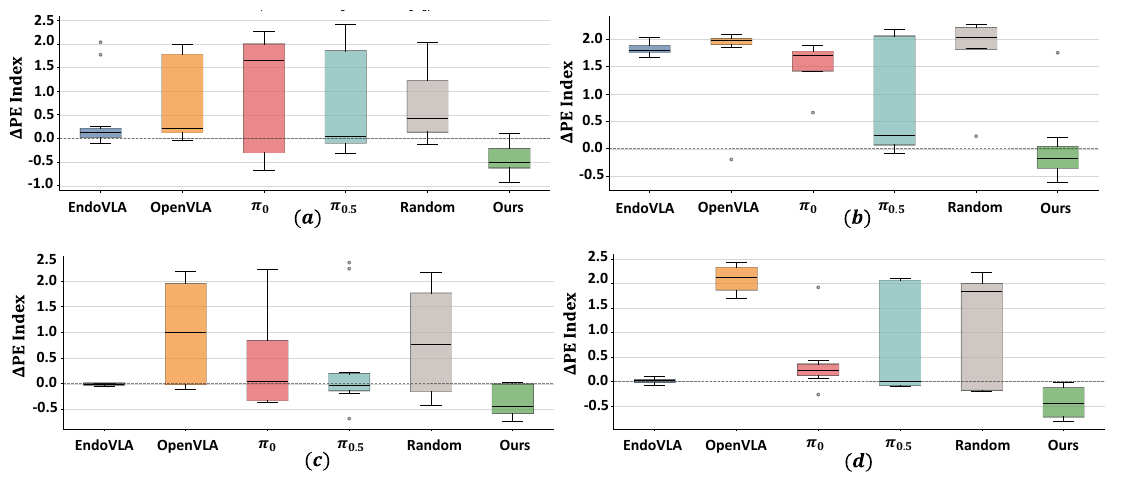}
\captionof{figure}{Centering experiment on R1 under (a) high-light dry, (b) high-light wet, (c) low-light dry and (d) low-light wet conditions. Box plots summarize the trial-level $\Delta PE$ index. The Ours label denotes Ours-A.}\label{fig:si-control}
\end{center}

Figure~\ref{fig:si-control} shows negative median $\Delta PE$ for Ours-A in all four conditions, indicating median net movement toward the image center. Several comparison methods show positive medians or distributions extending substantially above zero, with the pattern varying by appearance condition. Ours-A also exhibits positive outliers, especially in the high-light wet condition, showing variation across individual trajectories.

These results support the adapted policy's ability to reduce target-position error after the target is visible in the evaluated settings. They complement the acquisition experiments by examining camera control at the centering stage. The net-change statistic is interpreted together with the FT trial counts and trajectory examples, which report completion and temporal behavior, respectively.
\clearpage

\subsection{Find-and-track trajectories}

\begin{center}\includegraphics[width=\linewidth,height=525pt,keepaspectratio]{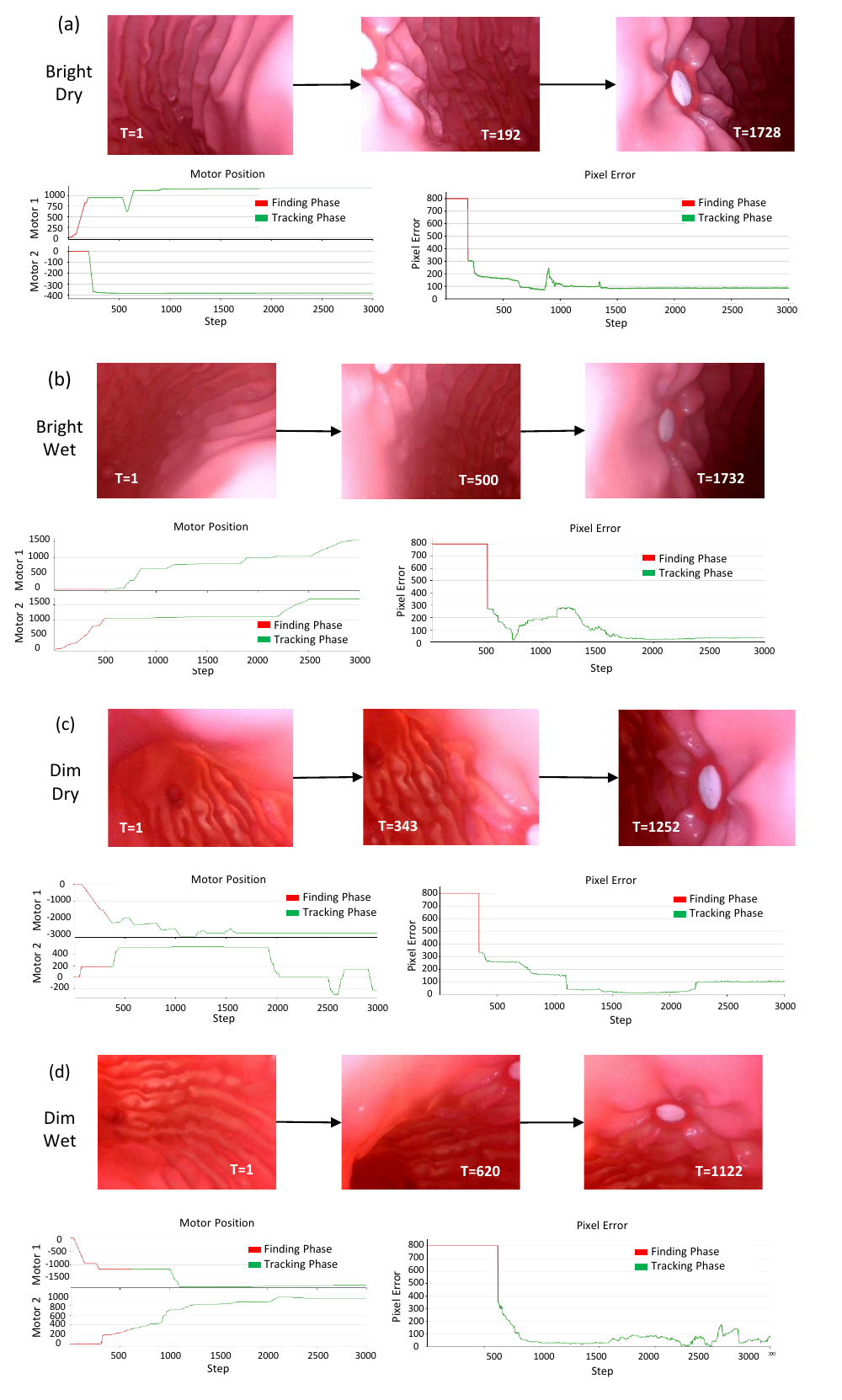}
\captionof{figure}{Successful Ours-A finding-and-tracking trajectories on R1 under (a) HD, (b) HW, (c) LD and (d) LW. Image sequences show search and approach; motor-position and pixel-error traces show the progression from finding to tracking.}\label{fig:si-traces}
\end{center}

Figure~\ref{fig:si-traces} connects the observed views with the commanded motion and target-position error in four successful trials. The traces begin at the out-of-view error convention and decrease after the target is acquired. Continued motor adjustments accompany the reduction in pixel error, showing how the policy progresses from search to local correction. The wet-condition examples contain further error fluctuations after acquisition, illustrating the corrective behavior required as the visible target position changes.

The figure demonstrates completed search-and-center trajectories across the four tested appearance conditions. These are individual successful rollouts; Table~\ref{tab:si-ft} supplies the corresponding condition-level FT counts of 8/9, 6/9, 8/9 and 5/9. Reading the examples alongside those counts connects the mechanism of a successful trial to its observed frequency over repeated attempts.
\clearpage

\section{Supplementary acquisition and transfer experiments}\label{sec:si-transfer}
\subsection{Acquisition across appearance conditions}
R1, R2 and training-phantom input interventions use the common acquisition protocol in Sec.~\ref{sec:si-protocols}: at least 100 consecutive visible frames within a 3,000-frame horizon, corresponding to five seconds within 150 seconds at 20 fps. After Find, composite FT requires the target-box center to enter the central rectangle, whose width and height are each 50\% of the image, with uninterrupted target visibility through the remaining trial. Center requires prior Find success followed by two continuous seconds inside the same rectangle with target visibility maintained. Each success rate uses all attempted trials as its denominator.

\setcounter{table}{7}
\renewcommand{\thetable}{S8a}
\captionof{table}{Finding outcomes on R1: successes/trials (percentage; 95\% Wilson interval).}\label{tab:si-find}
{\footnotesize\setlength{\tabcolsep}{4pt}\renewcommand{\arraystretch}{1.18}\begin{center}\begin{tabular}{p{\dimexpr 0.1400\linewidth-2\tabcolsep\relax}p{\dimexpr 0.2150\linewidth-2\tabcolsep\relax}p{\dimexpr 0.2150\linewidth-2\tabcolsep\relax}p{\dimexpr 0.2150\linewidth-2\tabcolsep\relax}p{\dimexpr 0.2150\linewidth-2\tabcolsep\relax}}\toprule\rowcolor{tableblue}

\color{white}\bfseries Method & \color{white}\bfseries HD & \color{white}\bfseries HW & \color{white}\bfseries LD & \color{white}\bfseries LW \\\midrule

Ours-A & 8/9 (88.9\%; 56.5-98.0) & 7/9 (77.8\%; 45.3-93.7) & 8/9 (88.9\%; 56.5-98.0) & 6/9 (66.7\%; 35.4-87.9) \\

MS & 5/9 (55.6\%; 26.7-81.1) & 3/9 (33.3\%; 12.1-64.6) & 3/9 (33.3\%; 12.1-64.6) & 1/9 (11.1\%; 2.0-43.5) \\

ES & 1/9 (11.1\%; 2.0-43.5) & 0/9 (0.0\%; 0.0-29.9) & 1/9 (11.1\%; 2.0-43.5) & 1/9 (11.1\%; 2.0-43.5) \\

$\pi_{0.5}$ & 2/9 (22.2\%; 6.3-54.7) & 1/9 (11.1\%; 2.0-43.5) & 0/9 (0.0\%; 0.0-29.9) & 0/9 (0.0\%; 0.0-29.9) \\

$\pi_0$ & 1/9 (11.1\%; 2.0-43.5) & 0/9 (0.0\%; 0.0-29.9) & 0/9 (0.0\%; 0.0-29.9) & 0/9 (0.0\%; 0.0-29.9) \\

OpenVLA & 0/9 (0.0\%; 0.0-29.9) & 0/9 (0.0\%; 0.0-29.9) & 0/9 (0.0\%; 0.0-29.9) & 0/9 (0.0\%; 0.0-29.9) \\

EndoVLA & 2/9 (22.2\%; 6.3-54.7) & 0/9 (0.0\%; 0.0-29.9) & 0/9 (0.0\%; 0.0-29.9) & 0/9 (0.0\%; 0.0-29.9) \\

\bottomrule\end{tabular}\end{center}}

\setcounter{table}{7}
\renewcommand{\thetable}{S8b}
\captionof{table}{Finding-and-tracking outcomes on R1: successes/trials (percentage; 95\% Wilson interval).}\label{tab:si-ft}
{\footnotesize\setlength{\tabcolsep}{4pt}\renewcommand{\arraystretch}{1.18}\begin{center}\begin{tabular}{p{\dimexpr 0.1400\linewidth-2\tabcolsep\relax}p{\dimexpr 0.2150\linewidth-2\tabcolsep\relax}p{\dimexpr 0.2150\linewidth-2\tabcolsep\relax}p{\dimexpr 0.2150\linewidth-2\tabcolsep\relax}p{\dimexpr 0.2150\linewidth-2\tabcolsep\relax}}\toprule\rowcolor{tableblue}

\color{white}\bfseries Method & \color{white}\bfseries HD & \color{white}\bfseries HW & \color{white}\bfseries LD & \color{white}\bfseries LW \\\midrule

Ours-A & 8/9 (88.9\%; 56.5-98.0) & 6/9 (66.7\%; 35.4-87.9) & 8/9 (88.9\%; 56.5-98.0) & 5/9 (55.6\%; 26.7-81.1) \\

MS & 5/9 (55.6\%; 26.7-81.1) & 3/9 (33.3\%; 12.1-64.6) & 3/9 (33.3\%; 12.1-64.6) & 1/9 (11.1\%; 2.0-43.5) \\

ES & 1/9 (11.1\%; 2.0-43.5) & 0/9 (0.0\%; 0.0-29.9) & 1/9 (11.1\%; 2.0-43.5) & 1/9 (11.1\%; 2.0-43.5) \\

$\pi_{0.5}$ & 0/9 (0.0\%; 0.0-29.9) & 0/9 (0.0\%; 0.0-29.9) & 0/9 (0.0\%; 0.0-29.9) & 0/9 (0.0\%; 0.0-29.9) \\

$\pi_0$ & 0/9 (0.0\%; 0.0-29.9) & 0/9 (0.0\%; 0.0-29.9) & 0/9 (0.0\%; 0.0-29.9) & 0/9 (0.0\%; 0.0-29.9) \\

OpenVLA & 0/9 (0.0\%; 0.0-29.9) & 0/9 (0.0\%; 0.0-29.9) & 0/9 (0.0\%; 0.0-29.9) & 0/9 (0.0\%; 0.0-29.9) \\

EndoVLA & 0/9 (0.0\%; 0.0-29.9) & 0/9 (0.0\%; 0.0-29.9) & 0/9 (0.0\%; 0.0-29.9) & 0/9 (0.0\%; 0.0-29.9) \\

\bottomrule\end{tabular}\end{center}}

Tables~\ref{tab:si-find} and \ref{tab:si-ft} show that Ours-A has the highest observed success count in each appearance condition for both endpoints. It achieves finding in 29/36 trials (80.6\%) and FT in 27/36 (75.0\%) across the four equally sampled conditions. The medical student achieves 12/36 for each endpoint and the engineering student 3/36. The other VLA comparisons have lower finding counts and FT counts of 0/9 per condition. Ours-A and all VLA baselines are fine-tuned on subset A.

Ours-A reaches 8/9 for both finding and FT in each dry condition. Wet views yield finding counts of 7/9 and 6/9 and FT counts of 6/9 and 5/9 under high and low light, respectively. The two-trial difference between total finding and FT arises in the wet conditions. This supports acquisition and completed FT across the evaluated appearance changes while identifying wet-view search and subsequent centering as practical targets for improvement. MS and ES each denote one student operator, from medicine and engineering, respectively; each completes nine trials per appearance condition.
\clearpage

\subsection{Transfer to unseen phantoms and ex-vivo tissue}

\setcounter{table}{8}
\renewcommand{\thetable}{S9}
\captionof{table}{Composite FT on the training and unseen phantoms using R1.}\label{tab:si-phantoms}
{\footnotesize\setlength{\tabcolsep}{4pt}\renewcommand{\arraystretch}{1.18}\begin{center}\begin{tabular}{p{\dimexpr 0.1600\linewidth-2\tabcolsep\relax}p{\dimexpr 0.2100\linewidth-2\tabcolsep\relax}p{\dimexpr 0.2100\linewidth-2\tabcolsep\relax}p{\dimexpr 0.2100\linewidth-2\tabcolsep\relax}p{\dimexpr 0.2100\linewidth-2\tabcolsep\relax}}\toprule\rowcolor{tableblue}

\color{white}\bfseries Phantom & \color{white}\bfseries HD & \color{white}\bfseries HW & \color{white}\bfseries LD & \color{white}\bfseries LW \\\midrule

Training & 8/9 (88.9\%; 56.5-98.0) & 6/9 (66.7\%; 35.4-87.9) & 8/9 (88.9\%; 56.5-98.0) & 5/9 (55.6\%; 26.7-81.1) \\

Unseen 1 & 6/9 (66.7\%; 35.4-87.9) & 4/9 (44.4\%; 18.9-73.3) & 6/9 (66.7\%; 35.4-87.9) & 3/9 (33.3\%; 12.1-64.6) \\

Unseen 2 & 5/9 (55.6\%; 26.7-81.1) & 5/9 (55.6\%; 26.7-81.1) & 7/9 (77.8\%; 45.3-93.7) & 3/9 (33.3\%; 12.1-64.6) \\

\bottomrule\end{tabular}\end{center}}

\setcounter{table}{9}
\renewcommand{\thetable}{S10}
\captionof{table}{Composite FT in ex-vivo anatomical regions using R1.}\label{tab:si-regions}
{\footnotesize\setlength{\tabcolsep}{4pt}\renewcommand{\arraystretch}{1.18}\begin{center}\begin{tabular}{p{\dimexpr 0.4000\linewidth-2\tabcolsep\relax}p{\dimexpr 0.6000\linewidth-2\tabcolsep\relax}}\toprule\rowcolor{tableblue}

\color{white}\bfseries Region & \color{white}\bfseries Successes / trials (\%; 95\% Wilson CI) \\\midrule

Greater curvature & 4/12 (33.3\%; 13.8-60.9) \\

Lesser curvature & 5/12 (41.7\%; 19.3-68.0) \\

Pyloric antrum & 6/12 (50.0\%; 25.4-74.6) \\

\bottomrule\end{tabular}\end{center}}

Table~\ref{tab:si-phantoms} reports 27/36 completed FT trials in the training phantom, 19/36 (52.8\%) in unseen phantom 1 and 20/36 (55.6\%) in unseen phantom 2. Both unseen phantoms contain successes in every tested appearance condition, with 3/9 in low-light wet views. The highest observed counts are 6/9 and 7/9 in dry views for phantoms 1 and 2, respectively. The results support transfer of the acquisition-and-centering behavior across the evaluated phantom geometries and identify the combined environment and appearance change as a more demanding setting.

Table~\ref{tab:si-regions} extends the R1 evaluation to ex-vivo tissue, with FT counts of 4/12 at the Greater Curvature, 5/12 at the Lesser Curvature and 6/12 at the Pyloric Antrum. Successful FT in all three regions supports endoscopic target revisitation in the evaluated tissue preparation. Together, Tables~\ref{tab:si-phantoms} and \ref{tab:si-regions} provide environment-transfer evidence on R1, complementing the R2 cross-embodiment evaluation reported in the main manuscript.

\subsection{Trial-level uncertainty}
Tables~\ref{tab:si-find}--\ref{tab:si-regions} report two-sided 95\% Wilson intervals. For $k$ successes in $n$ trials, $\hat p=k/n$ and $z=1.959964$, the interval is
\begin{equation}
\frac{\hat p+z^2/(2n)\;\pm\;z\sqrt{\hat p(1-\hat p)/n+z^2/(4n^2)}}{1+z^2/n}.
\label{eq:si-wilson}
\end{equation}
The intervals describe binomial counting uncertainty at the trial level. Repeated trials share a phantom or tissue preparation, so interpretation remains specific to the evaluated conditions and specimens. The tables use nine trials per appearance condition and twelve per ex-vivo anatomical region.

\clearpage
\subsection{Failure patterns across conditions and task stages}\label{sec:si-failure-strata}
This descriptive analysis reorganizes the physical trial counts in Tables~\ref{tab:si-find}--\ref{tab:si-phantoms} and the main-text R2 results. It uses the same attempts and success criteria. For $N$ attempts, $F$ Find successes and $C$ Center successes, the mutually exclusive outcomes are $N-F$ acquisition failures, $F-C$ successful acquisitions without Center, and $C$ completed Center trials. The conditional fraction $(F-C)/F$ describes centering failure among acquired targets. R1 FT additionally includes retention, so its post-Find failures encompass centering and subsequent visibility.

\textbf{Appearance-associated failures on R1.}
Table~\ref{tab:si-failure-appearance} pools the two equally sampled lighting conditions within each phantom. Wet views yield more FT failures in all three phantoms: 7 versus 2 in the training phantom, 11 versus 6 in unseen phantom 1, and 10 versus 6 in unseen phantom 2, each comparison using 18 attempts per appearance group. Across the 108 distinct trials, the failure fractions are 28/54 (51.9\%) for wet and 14/54 (25.9\%) for dry conditions, a descriptive difference of 25.9 percentage points. The pooled difference therefore reflects a pattern present in each phantom.

\setcounter{table}{10}\renewcommand{\thetable}{S11a}
\captionof{table}{Ours-A FT failures on R1, stratified by phantom or lighting. Entries are failures/attempts (\%). The lighting rows regroup the same trials used in the phantom rows.}\label{tab:si-failure-appearance}
{\small\setlength{\tabcolsep}{5pt}\renewcommand{\arraystretch}{1.22}\begin{center}\begin{tabularx}{\linewidth}{>{\raggedright\arraybackslash}Xrr}\toprule\rowcolor{tableblue}
\color{white}\bfseries Stratum & \color{white}\bfseries Dry & \color{white}\bfseries Wet \\\midrule
Training phantom & 2/18 (11.1) & 7/18 (38.9) \\
Unseen phantom 1 & 6/18 (33.3) & 11/18 (61.1) \\
Unseen phantom 2 & 6/18 (33.3) & 10/18 (55.6) \\
\textbf{All three phantoms} & \textbf{14/54 (25.9)} & \textbf{28/54 (51.9)} \\\midrule
High light, three phantoms & 8/27 (29.6) & 12/27 (44.4) \\
Low light, three phantoms & 6/27 (22.2) & 16/27 (59.3) \\
\bottomrule\end{tabularx}\end{center}}

Stratifying by lighting also yields more wet-view failures: 12/27 versus 8/27 under high light and 16/27 versus 6/27 under low light. Low-light wet views have the largest pooled failure count, whereas low-light dry views have the smallest. Thus, the observed boundary concerns wet appearance and its combination with lighting; low illumination alone has a different pattern. Repeated trials share three phantoms, and the contrasts describe these evaluated settings.

The training-phantom counts further locate the failure stage. Dry conditions contain two acquisition failures in 18 attempts and successful FT in all 16 acquired-target trials. Wet conditions contain five acquisition failures in 18 attempts, followed by two FT failures among the 13 acquired-target trials. Wet appearance therefore accompanies difficulty both before acquisition and during subsequent centering or retention. The droplet and bubble examples in main Fig.~12 illustrate plausible visual-discrimination challenges within this broader pattern. Their role is qualitative: assigning a frequency or causal contribution to either mode requires per-trial mode annotations.

\textbf{Evidence boundary for initial pose.}
The archived controls record tip position, orientation and motor-position proxies. Estimating a pose-dependent failure boundary additionally requires an unambiguous join between each initial configuration and its final reviewed outcome, together with a common spatial reference. The present analysis uses verified condition-level outcomes and leaves pose-conditioned failure probabilities unestimated. A trial-indexed pose/outcome evaluation would separate difficult starting configurations from appearance and instruction effects.

\clearpage
\textbf{Instruction conditions and post-acquisition control on R2.}
Table~\ref{tab:si-failure-stages} decomposes the R2 outcomes using the common Find-before-Center rule. On standard exA--C trials, two attempts fail acquisition and eight acquired targets fail Center. Thus, 8/10 of the attempts failing Center reach the target first, and the conditional centering-failure fraction is 8/28 (28.6\%). The corresponding fractions are 2/10 on exA and 3/9 on each of exB and exC. These identify remaining post-acquisition difficulty in the evaluated organ--target configurations; specimen, target appearance and anatomical region vary together.

\renewcommand{\thetable}{S11b}
\captionof{table}{R2 failure-stage decomposition. Counts in the three outcome columns sum to $N$. The final column uses successful acquisitions $F$ as its denominator. All rows use Ours-B except the specified Ours-A comparison. The pooled standard row repeats the first three rows.}\label{tab:si-failure-stages}
{\footnotesize\setlength{\tabcolsep}{3pt}\renewcommand{\arraystretch}{1.25}\begin{center}\begin{tabularx}{\linewidth}{>{\raggedright\arraybackslash}Xr>{\centering\arraybackslash}p{47pt}>{\centering\arraybackslash}p{65pt}r>{\centering\arraybackslash}p{76pt}}\toprule\rowcolor{tableblue}
\color{white}\bfseries Condition & \color{white}\bfseries $N$ & \color{white}\bfseries Find failure & \color{white}\bfseries Find without Center & \color{white}\bfseries Center & \color{white}\bfseries $(F-C)/F$ (\%) \\\midrule
exA standard & 10 & 0 & 2 & 8 & 2/10 (20.0) \\
exB standard & 10 & 1 & 3 & 6 & 3/9 (33.3) \\
exC standard & 10 & 1 & 3 & 6 & 3/9 (33.3) \\
\textbf{exA--C standard} & \textbf{30} & \textbf{2} & \textbf{8} & \textbf{20} & \textbf{8/28 (28.6)} \\\midrule
exA full instruction & 10 & 0 & 2 & 8 & 2/10 (20.0) \\
exB unseen correct wording & 10 & 0 & 4 & 6 & 4/10 (40.0) \\
exC empty instruction & 10 & 7 & 3 & 0 & 3/3 (100.0) \\
exA wrong-region request & 10 & 8 & 0 & 2 & 0/2 (0.0) \\
exC after occlusion & 10 & 1 & 3 & 6 & 3/9 (33.3) \\\midrule
exD--F Ours-A & 30 & 18 & 5 & 7 & 5/12 (41.7) \\
exD--F Ours-B & 30 & 3 & 8 & 19 & 8/27 (29.6) \\
\bottomrule\end{tabularx}\end{center}}

On exC, empty instructions yield seven acquisition failures, compared with one under the standard prompt. On exA, the wrong-region request yields eight reference-target acquisition failures, compared with zero in either the standard or full-instruction group. The requested wrong region is target-free; these outcomes measure the reference-task consequence of an altered request. Unseen correct wording on exB preserves acquisition in all ten attempts, with four subsequent Center failures. These within-organ observations associate instruction content with the acquisition stage, while correct semantic guidance still leaves a local-control challenge.

Post-occlusion recovery on exC gives one acquisition failure and three post-acquisition Center failures, matching the aggregate standard exC outcomes. The observed counts support recovery with remaining precision failures; they provide neither an excess-failure estimate attributable to occlusion nor a test of equivalence.

For exD--F, acquisition failures decrease from 18/30 with Ours-A to 3/30 with Ours-B. Conditional Center failure is 5/12 (41.7\%) versus 8/27 (29.6\%) among acquired targets. The larger absolute post-acquisition failure count for Ours-B occurs alongside 15 additional acquisitions; the conditional denominator clarifies that difference. These post-acquisition subsets contain different successful trials, so the fractions describe each system's observed outcomes. Together, the strata prioritize robust search under wet appearance, informative region--target requests, and precise correction after acquisition. They locate observed difficulties without attributing them to a particular initial pose or inference latency.

\end{document}